\documentclass{article}

    \PassOptionsToPackage{numbers, compress}{natbib}
\usepackage[preprint]{neurips_2024}

\usepackage[utf8]{inputenc} % allow utf-8 input
\usepackage[T1]{fontenc}    % use 8-bit T1 fonts
\usepackage{hyperref}       % hyperlinks
\usepackage{url}            % simple URL typesetting
\usepackage{booktabs}       % professional-quality tables
\usepackage{amsfonts}       % blackboard math symbols
\usepackage{nicefrac}       % compact symbols for 1/2, etc.
\usepackage{microtype}      % microtypography
\usepackage{xcolor}         % colors

\usepackage{xspace}
\usepackage{tcolorbox}
\usepackage{algorithm}  % 使用您提供的 algorithm.sty
\usepackage{algorithmic}  % 使用您提供的 algorithmic.sty
\usepackage{amsmath}

\usepackage{booktabs}  % 用于 \toprule, \midrule, \bottomrule
\usepackage{multirow}  % 用于 \multirow
\usepackage{makecell}  % 用于 \makecell
\usepackage{colortbl}  % 用于 \rowcolors 和 \cellcolor
\usepackage{xcolor}    % 用于颜色定义
\usepackage{subfigure}  % 推荐使用这个，更现代的包

\newcommand{\modelname}{GraphTeam\xspace}

\definecolor{shallowpurple}{RGB}{224,223,255}
\definecolor{shallowred}{RGB}{242,204,208}
\definecolor{shallowgreen}{RGB}{204,242,208}  % #CCF2D0

\definecolor{shallowblue}{RGB}{204,208,242}   % #CCD0F2

\definecolor{shallowgray}{RGB}{237,240,246}
\newtcolorbox{shallowpurple}[1][]{
	width=\columnwidth,
	colback = shallowpurple, 
	colframe = shallowpurple, 
	boxsep=0pt,left=10pt,right=10pt,top=5pt,bottom=5pt,
	fontupper=\linespread{0.9}\selectfont,
	title=#1}

\newtcolorbox{shallowred}[1][]{
	width=\columnwidth,
	colback = shallowred, 
	colframe = shallowred, 
	boxsep=0pt,left=10pt,right=10pt,top=5pt,bottom=5pt,
	fontupper=\linespread{0.9}\selectfont,
	title=#1}

\newtcolorbox{shallowgreen}[1][]{
	width=\columnwidth,
	colback = shallowgreen, 
	colframe = shallowgreen, 
	boxsep=0pt,left=10pt,right=10pt,top=5pt,bottom=5pt,
	fontupper=\linespread{0.9}\selectfont,
	title=#1}

\newtcolorbox{shallowblue}[1][]{
	width=\columnwidth,
	colback = shallowblue, 
	colframe = shallowblue, 
	boxsep=0pt,left=10pt,right=10pt,top=5pt,bottom=5pt,
	fontupper=\linespread{0.9}\selectfont,
	title=#1}

\newtcolorbox{prompt}[1][]{
	width=\columnwidth,
	colback = shallowgray, 
	colframe = shallowgray, 
	boxsep=0pt,left=10pt,right=10pt,top=5pt,bottom=5pt,
	fontupper=\linespread{0.9}\selectfont,
	title=#1}

\title{\modelname: Facilitating Large Language Model-based Graph Reasoning via Multi-Agent Collaboration}

\author{Xin Li$^{1*}$, Qizhi Chu$^{1*}$, Yubin Chen$^{2*}$, Yang Liu$^{1}$, Yaoqi Liu$^{1}$, 
\\\textbf{Zekai Yu$^{1}$, Weize Chen$^{3}$, Chen Qian$^{3}$, Chuan Shi$^{1}$, Cheng Yang$^{1\dagger}$}\\
\\
$^1$ Beijing University of Posts and Telecommunications\\
$^2$ The Chinese University of Hong Kong\\
$^3$ Tsinghua University,\\
\\
\texttt{lixin4sky@bupt.edu.cn},\quad \texttt{chuqizhi@bupt.edu.cn}, \\
\texttt{YubinChen@link.cuhk.edu.hk}, \quad
\texttt{yangcheng@bupt.edu.cn} \\
}

\begin{document}

\maketitle

% {\centering\large\color{black} WORK IN PROGRESS - DRAFT VERSION\par}
% \vspace{1em}

\begin{abstract}
    Graphs are widely used for modeling relational data in real-world scenarios, such as social networks and urban computing. While large language models (LLMs) have achieved strong performance in many areas, existing LLM-based graph analysis approaches either integrate graph neural networks (GNNs) for specific machine learning tasks (\textit{e.g.,} node classification), limiting their transferability, or rely solely on LLMs' internal reasoning ability, resulting in suboptimal performance. To address these limitations, we take advantage of recent advances in LLM-based agents, which have shown capabilities of utilizing external knowledge or tools for problem solving. By simulating human problem-solving strategies such as analogy and collaboration, we propose a multi-agent system based on LLMs named \modelname, for graph analysis. \modelname consists of five LLM-based agents from three modules, and the agents with different specialities can collaborate with each other to address complex problems. Specifically, (1) \textit{input-output normalization module}: the question agent extracts and refines four key arguments (\textit{e.g.,} graph type and output format) from the original question, facilitating the problem understanding, and the answer agent organizes the results to meet the output requirement; (2) \textit{external knowledge retrieval module}: we first build a knowledge base consisting of relevant documentation and experience information, and then the search agent retrieves the most relevant entries from the knowledge base for each question. (3) \textit{problem-solving module}: given the retrieved information from search agent, the coding agent uses established algorithms via programming to generate solutions, and in case the coding agent does not work, the reasoning agent will directly compute the results without programming. Extensive experiments on six graph analysis benchmarks demonstrate that \modelname achieves state-of-the-art performance with an average 25.85\% improvement over the best baseline in terms of accuracy. 
\end{abstract}

\section{Introduction}
% \input{GraphTeam/introduction}
% \section{Introduction}
% a story of cognitive process of problem solving human

\begin{figure}[ht]
    \centering
    \small
    \begin{subfigure}
        \centering
        \label{fig:simplified_modules}
        \includegraphics[width=0.5\linewidth]{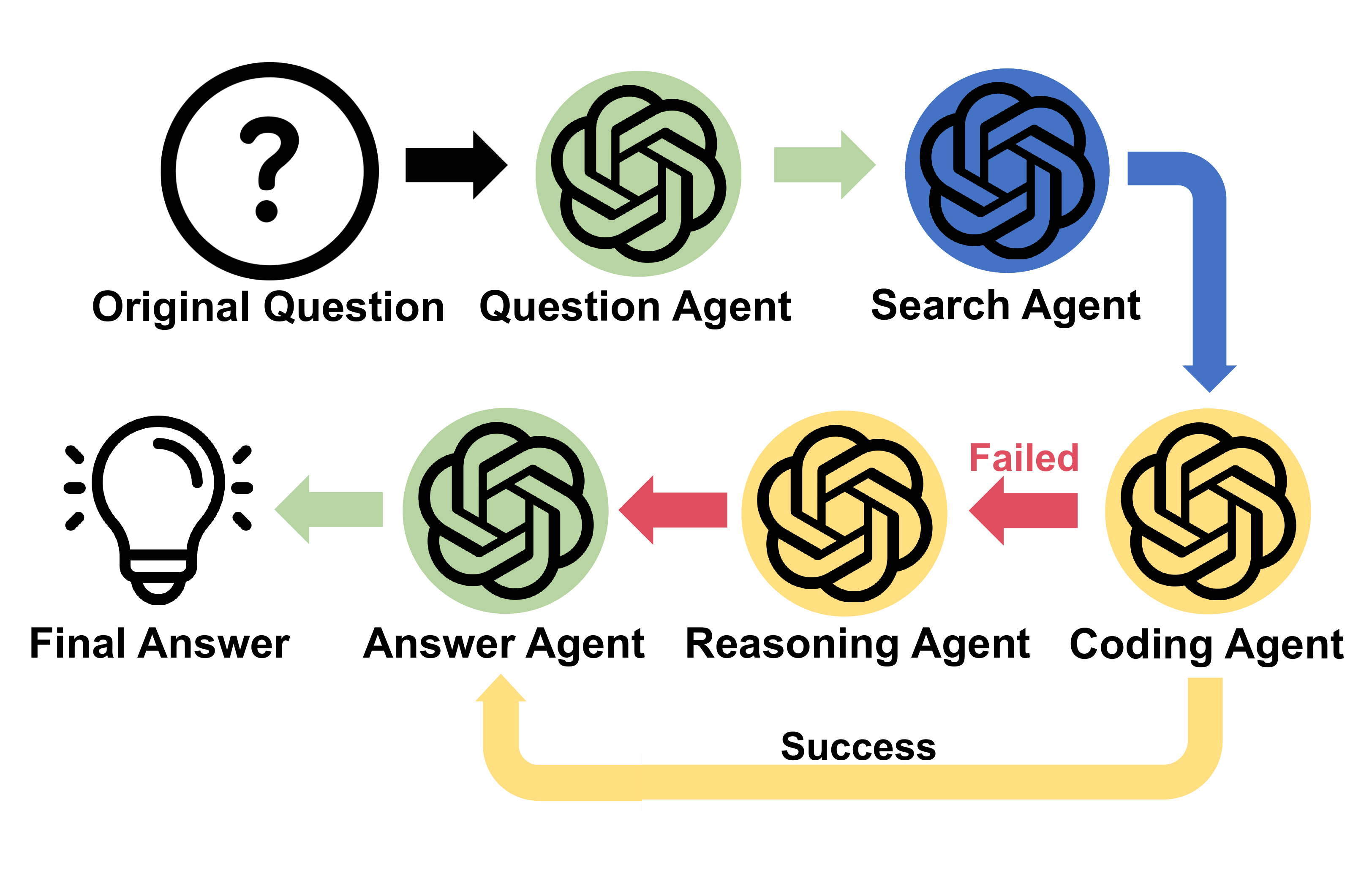}
        % \caption{The simplified modules}
    \end{subfigure}
    \hfill
    \begin{subfigure}
        \centering
        \label{fig:radar}
        \includegraphics[width=0.45\linewidth]{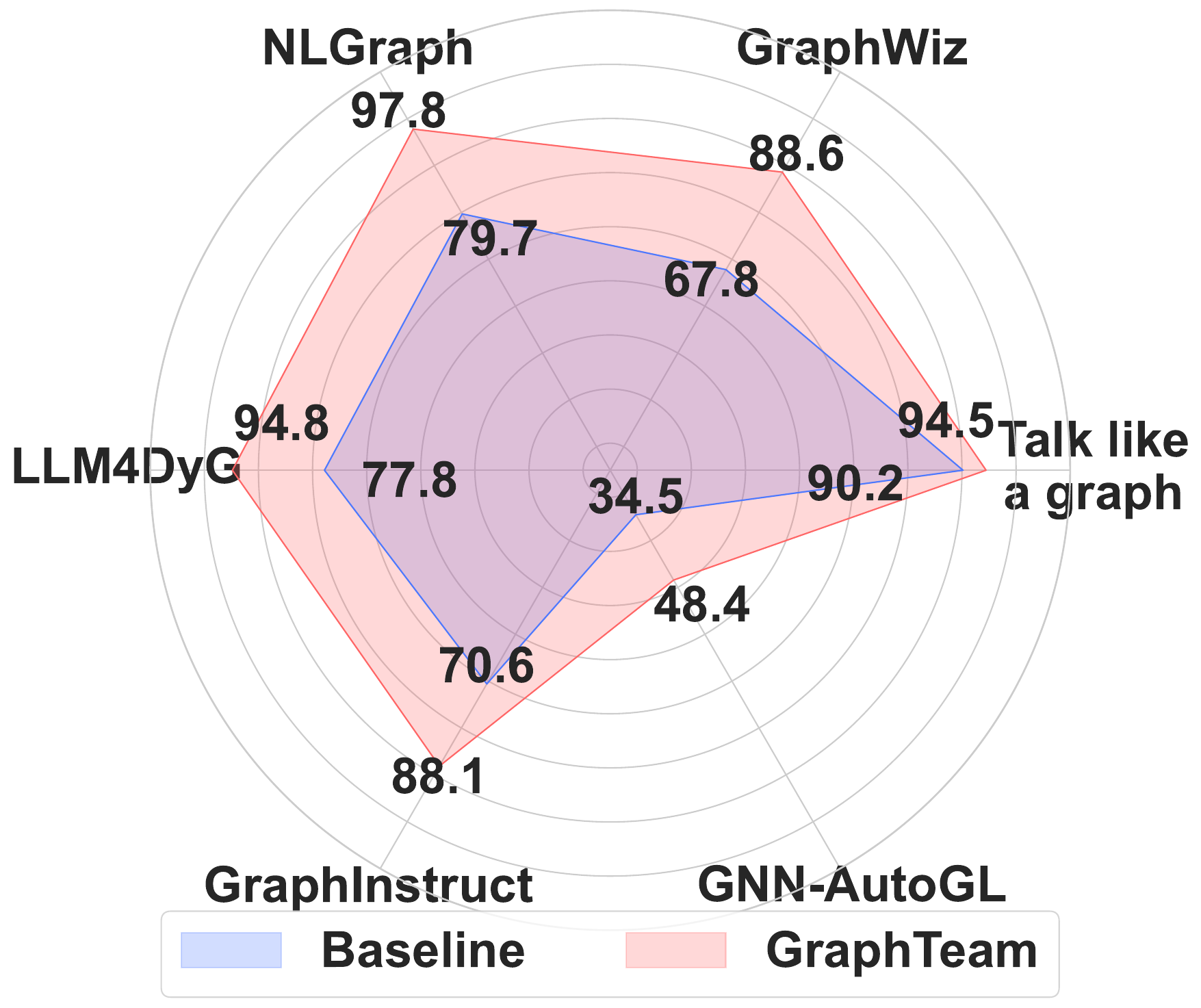}
        % \caption{Overall results}
        %\vspace{-1.8em}
    \end{subfigure}
    
    \caption{The overall pipeline of our multi-agent system \modelname (left), and the comparison between \modelname and the best baseline on six benchmarks (right).}
    \label{fig:simplified_modules_and_overall_performance}
    \vspace{-2em}
\end{figure}

Social networks~\cite{social_networks}, urban computing~\cite{zheng2014urban}, and various other fields make extensive use of graph data and related analytical methods. Given the promising problem-solving abilities demonstrated by large language models (LLMs)~\cite{gpt3, gemini, claude3} across different domains, researchers are now exploring the feasibility of using LLMs to reason over graphs~\cite{ graphgpt,transformer_reasoning, GUNDAM, zhang2024graphtranslator}.

Existing LLM-based graph reasoning methods mainly fall into two categories. The first category~\cite{graphllm, liu2023molca,  qin2023disentangled, liu2024git} usually combines LLMs with graph neural networks (GNNs) for modeling. A typical paradigm is to first encode the input graph with a GNN, and then feed the encodings to an LLM as language token embeddings. However, this line of work is specialized for graph machine learning tasks, especially node classification, and thus has limited transferability. The second category~\cite{talk_like_a_graph, Graph_LLM} usually flattens graphs into textual descriptions, and purely relies on LLMs for analyzing~\cite{liu2023evaluating, huang2023can, graphtext, relm, Graph_LLM}. Strategies like in-context learning~\cite{gpt3} and chain-of-thought reasoning~\cite{chain_of_thought} are widely used in these methods to improve performance. However, these methods analyze graphs by step-by-step reasoning on raw inputs without utilizing any external knowledge or tools, severely limiting their performance.

Fortunately, LLMs have been shown to be capable of using external knowledge or tools for solving problems, known as a key feature of LLM-based agents~\cite{fudan_nlp_agent_survey, ruc_agent_survey}. Moreover, inspired by the division of labor and cooperation mechanisms in human society, researchers propose LLM-based multi-agent systems~\cite{ChatDev, MetaGPT, autogen, AgentVerse}, where agents with different specialities can collaborate with each other to address complex problems. But previous multi-agent frameworks are designed for general purposes, and have few knowledge relevant to graph 
reasoning.

Inspired by human problem-solving processes~\cite{how_to_solve_it, information_problem_solving, bscs_5e, investigating_learning_outcomes}, we present \modelname, a multi-agent system with three core modules specialized for graph reasoning: (1) \textit{Input-output normalization module.} In human problem solving, clearly understanding and formalizing the problem is the first step~\cite{how_to_solve_it, bscs_5e}, and answering the question in required formats is also important. Therefore, we introduce two different LLM-based agents: the question agent extracts and refines key arguments from the original question to help the system better understand the problem; the answer agent reformats the results to meet the specified output requirement. (2) \textit{External knowledge retrieval module.} In tackling complex problems, humans often seek help from external sources~\cite{information_problem_solving} or draw analogies to previously solved ones~\cite{bscs_5e}. Therefore, we first build a knowledge base that includes relevant documentation and experience information. Then we introduce a search agent to retrieve the most relevant entries from this knowledge base to assist the downstream solving process. (3) \textit{Problem-solving module.} Given the retrieved information, we introduce a coding agent that uses established algorithms via programming to generate solutions. If the coding agent fails, a reasoning agent directly infers the solution without relying on programming. The overall pipeline of \modelname is shown in Figure~\ref{fig:simplified_modules_and_overall_performance}. We evaluate \modelname through extensive experiments across six graph reasoning benchmarks. These benchmarks cover a variety of graph tasks, ranging from basic graph theory to the deployment of GNNs. The experimental results and ablation studies highlight the effectiveness of \modelname as well as the necessity of each module. In summary, our contributions are three-fold:

$\bullet$ Different from previous paradigms that utilize LLMs for graph reasoning, we propose to facilitate the solving process via the collaboration of multiple LLM-based agents.

$\bullet$ Inspired by human problem-solving processes, we propose \modelname, a carefully designed framework consisting of five agents across three functional groups: input-output normalization, external knowledge retrieval, and problem-solving modules. The agents collaborate with a well-defined division of labor for graph reasoning.

$\bullet$ Extensive experiments show that \modelname achieves state-of-the-art (SOTA) performance across all six graph reasoning benchmarks, with an average accuracy improvement of 15.3\% over the best baseline. 

\section{Related Work}
% \input{GraphTeam/related_work}
% \section{Related Work}
\subsection{LLM-based Graph Reasoning}

% \subsubsection{Benchmarks}
To evaluate the ability of LLMs in solving graph reasoning problems, researchers have developed various benchmarks~\cite{GPT4Graph, talk_like_a_graph, graphtmi, graphwiz, prograph}. Among them, NLGraph~\cite{nlgraph} is one of the first benchmark for assessing LLMs' graph analysis capabilities, focusing on basic graph theory reasoning and simple GNN computations. Graphwiz~\cite{graphwiz} evaluates LLMs on graph problems of varying complexity, including NP-complete ones. GraphInstruct~\cite{graphinstruct} includes 21 graph analysis tasks covering node-level, node-pair-level, and graph-level problems. Talk like a Graph~\cite{talk_like_a_graph} measures basic graph understanding abilities, while LLM4DyG~\cite{LLM4DyG} focuses on dynamic graph tasks related to temporal and spatial information. In our experiments, we employ the above five benchmarks, and also construct an additional one to evaluate the ability of deploying GNNs with AutoGL~\cite{AutoGL}.

There is a rising trend of research to utilize LLMs for various graph reasoning tasks~\cite{li2024surveygraphmeetslarge, jin2024largelanguagemodelsgraphs}. Here we focus on the most relevant methods that employ LLMs as predictors~\cite{graphllm, liu2023molca, qin2023disentangled, liu2024git}. Related work can be roughly divided into two categories. The first category typically integrates LLMs with graph neural network (GNNs), and is specialized for graph machine learning tasks such as node classification~\cite{graphgpt, zhang2024graphtranslator}. For example, GraphGPT~\cite{graphgpt} employs instruction tuning to fine-tune a GNN-enhanced LLM for node classification. GraphTranslator~\cite{zhang2024graphtranslator} encodes nodes in a graph by a GNN, and then project node embeddings into the embedding space of LLM tokens for zero-shot classification. However, these methods are designed and optimized for a specific task, and cannot generalize well to the diverse reasoning tasks in the above benchmarks. The second category usually flatten graphs into textual descriptions of adjacency lists, and then rely on LLM itself to analyze graphs with strategies like in-context learning~\cite{gpt3}, chain-of-thought reasoning~\cite{chain_of_thought} for inference~\cite{liu2023evaluating, huang2023can, graphtext, relm, Graph_LLM}. For example, GUNDAM~\cite{GUNDAM} collects natural language-based question-answer pairs of graph analysis problems to fine-tune LLMs, and performs step-by-step reasoning for inference. However, recent studies have shown that the reasoning depths of current LLMs are still
shallow~\cite{parmar2024causalchaos, li2024faithful}, and will probably fail for problems that require complex reasoning. In contrast, we encourage LLMs to leverage external knowledge and tools, instead of performing step-by-step reasoning on raw inputs.

\subsection{Multi-Agent Collaboration of LLMs}

Autonomous agents aim to solve problems with self-directed planning and actions. Recently, researchers introduce the concept of LLM-based agent that is capable of understanding natural language instruction, interacting with humans, perceiving external environments, and performing various actions~\cite{bang2023multitask, li2023modelscope, fudan_nlp_agent_survey, ruc_agent_survey}. Moreover, inspired by the collaboration of human in solving problems, researchers introduce LLM-based multi-agent systems, where agents collaborate with a well-defined division of labor. 
For example, ChatDev~\cite{ChatDev} is a virtual software company where LLM-based agents are responsible for different software development tasks (such as user interface design or software testing) for collaboration. MetaGPT~\cite{MetaGPT} is a meta-programming framework that simulates human workflows. With proper developing pipeline, MetaGPT can generate effective programming solutions. AgentVerse~\cite{AgentVerse} is a multi-agent framework inspired by human collaboration dynamics. It involves four iterative stages of expert recruitment, decision discussion, action and evaluation. AutoGen~\cite{autogen} is another multi-agent framework for general purposes. Through tool invocation and code generation, AutoGen can accomplish a wide variety of tasks. MACRec~\cite{MACRec} is a novel framework designed to enhance recommendation
systems through multi-agent collaboration. However, these multi-agent frameworks are not specialized for graph reasoning, \textit{e.g.,} they have few knowledge of relevant Python libraries and no experience base of similar problems.

\section{Methodology}
% \input{GraphTeam/methodology}
% \section{Methodology}

We start by presenting the overall framework of \modelname, and then present the details of each agent in the system. 
\subsection{Framework Overview}
As shown in Figure~\ref{fig:modules}, the entire framework of \modelname consists of five agents across three functional groups:

\begin{figure*}
    \centering
    \includegraphics[width=1.0\linewidth]{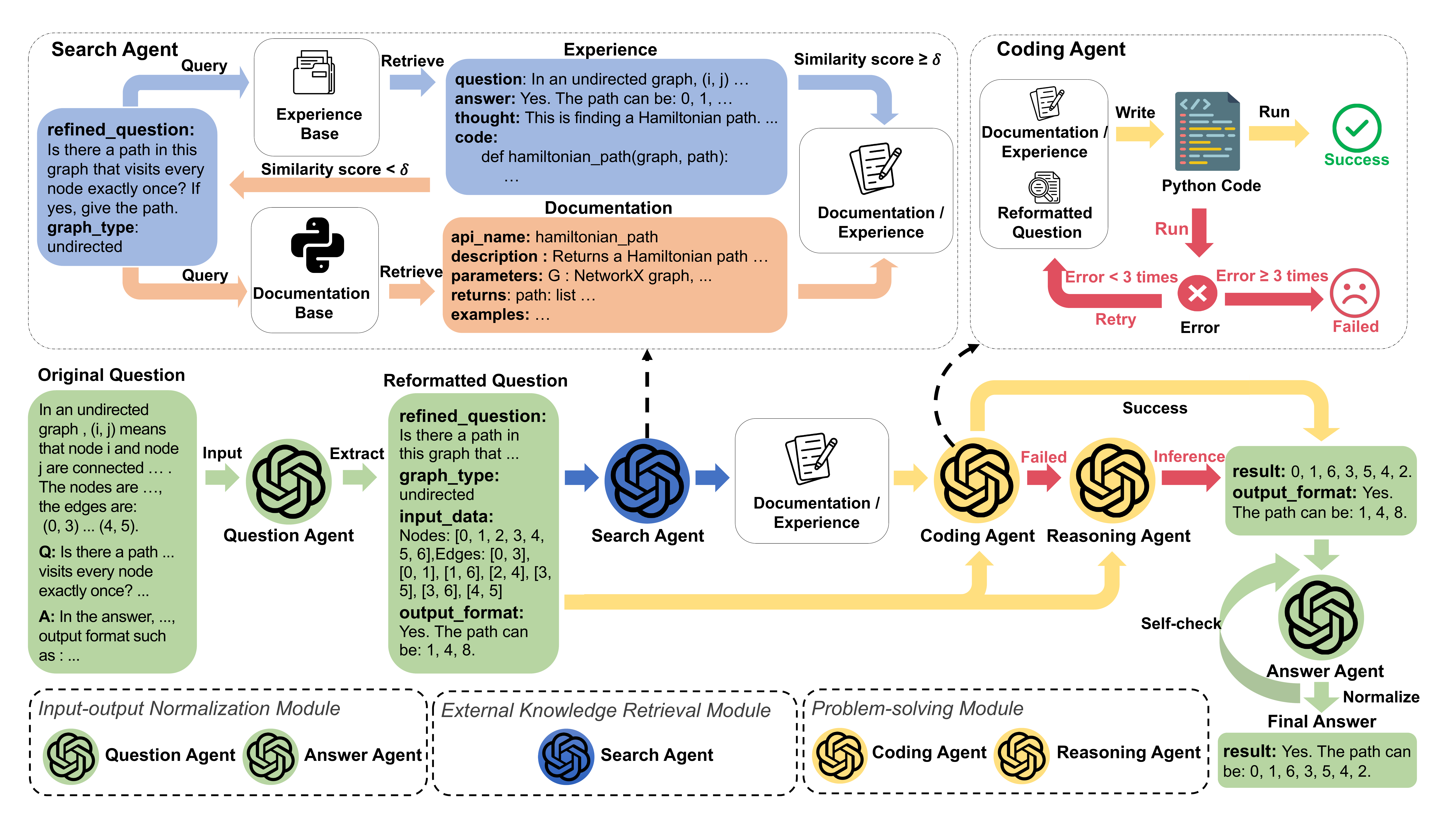}
    \vspace{-2em}
    \caption{The overall framework of \modelname, which includes five agents across three functional groups. {Firstly, the question agent refines the original question by extracting key arguments. Then the search agent queries knowledge bases with the reformatted question to retrieve relevant experience and documentation. Subsequently, the coding agent attempts to solve the problem via programming. If the coding agent fails after several retries, the reasoning agent will take over and directly solve the question without programming. Finally, the solutions from either the coding or reasoning agents are then passed to the answer agent for format standardization.}}
    \label{fig:modules}
    \vspace{-1em}
\end{figure*}

\textbf{Input-Output Normalization}: This functional group aims to refine key information from input questions, and map the results to desired output formats. Specifically, the question agent is responsible for processing the original problem descriptions, extracting key arguments to assist the system in solving it. The answer agent is responsible for transferring the formats of calculated results, and conduct self-checking to ensure the consistency with problem requirements.

\textbf{External Knowledge Retrieval}: This functional group only includes the search agent to extract relevant information from external knowledge base. Specifically, we first build a knowledge base for graph reasoning based on documentation of popular Python libraries and experiences of previous problem-solving processes. With problems refined by the upstream question agent as queries, the search agent then retrieves information beneficial to downstream problem solving from the knowledge base.

\textbf{Problem Solving}: This functional group calculates the results based on the refined question of question agent and the extracted knowledge of search agent. Specifically, the coding agent attempts to solve the problem via programming. If the coding agent fails after several retries, the reasoning agent will solve the problem directly without programming.

\subsection{Question Agent}

Abstraction is a preliminary step for human beings to solve problems~\cite{how_to_solve_it}. Inspired by this, we introduce a question agent to extract and refine four key arguments from the original problem descriptions. Formally, given a graph reasoning problem with its original description denoted as $q$, we employ an LLM denoted as $\operatorname{Question}(q)$ to generate refined question $q_r$, graph type $q_t$, input graph $q_g$ and output format $q_f$:
\begin{equation}
    (q_r,q_t,q_g,q_f)\leftarrow \operatorname{Question}(q),
\end{equation}
where $q_r$ is a condensed version of original instruction, $q_t$ indicates the graph type (\textit{e.g.,} directed, undirected, weighted, etc.), $q_g$ contains the graph to be processed, and $q_f$ specifies how the answer should be presented (\textit{e.g.,} numerical value, list, etc.). By explicitly identifying these key information, downstream agents can better understand a task and its requirements, thereby enhancing the overall reasoning capability. An example of question agent is shown below:

\begin{tcolorbox}[colback=gray!10, colframe=black, rounded corners, boxrule=1.5pt, fontupper=\normalsize, left=2mm, right=2mm, top=1mm, bottom=1mm]
    % \vspace{1em}
    \textbf{Original Question} \newline
    In an undirected graph ..., the edges are: (0,3) ... (4,5).\newline
    Q: Is there a path ... visits every node exactly once? ...\newline
    A: In the answer, ..., Output format such as : ... \newline
    $\textbf{Reformatted Question}$ \newline
    $\text{refined\_question}$: Is there a path in this graph that ...\newline
    $\text{graph\_type}$: undirected \newline
    $\text{input\_data}$: {Nodes: [0, 1, 2, 3, 4, 5, 6], \newline
    Edges: [0, 3], [0, 1], [1, 6], [2, 4], [3, 5], [3, 6], [4, 5]} \newline
    $\text{output\_format}$: Yes. The path can be: 1,4,8.
\end{tcolorbox}

\subsection{Search Agent}

During the solving process of a hard problem, it is common for human beings to seek help from external sources~\cite{information_problem_solving} or make analogy to a previously solved one~\cite{bscs_5e}. Therefore, we first build a knowledge base with documentation and experience information, and then ask the search agent to retrieve relevant entries from the knowledge base for each question.

\subsubsection{Knowledge Base Construction} The knowledge base has two parts, \textit{i.e.,} documentation base $\mathcal{K}_{doc}$ and experience base $\mathcal{K}_{exp}$. Examples of documentation and experience entries are shown below:

\begin{tcolorbox}[colback=gray!10, colframe=black, rounded corners, boxrule=1.5pt, fontupper=\normalsize, left=2mm, right=2mm, top=1mm, bottom=1mm]
    \textbf{{Documentation Information}} \newline
    \text{api\_name}: hamiltonian\_path \newline
    \text{description}: Returns a Hamiltonian path ... \newline
    \text{parameters}: 
    {
    G : NetworkX graph, ...
    } \newline
    \text{returns}: {
    path: list...
    % which form a Hamiltonian path in G."
    } \newline
    \text{examples}:
    ...
    % \begin{verbatim}
    % import NetworkX as nx
    % ...
    % nx.tournament.hamiltonian_path(G)
    % \end{verbatim}
    
\end{tcolorbox}
% \end{minipage}

%\subsubsection{Experience}

% \begin{minipage}[t]{0.46\textwidth}
\begin{tcolorbox}[colback=gray!10, colframe=black, rounded corners, boxrule=1.5pt, fontupper=\normalsize, left=2mm, right=2mm, top=1mm, bottom=1mm]
    \textbf{{Experience Information}} \newline
    \text{question}: In an undirected graph, ..., the edges are: (0,1) ... (5,6) 
    Q: Is there a path in this graph ... \newline
    \text{answer}: Yes. The path can be: 0, 1, 6, 5. \newline
    % \text{meta information}:
    % \newline
    % \{ \newline
        %\text{refined\_question}: Is there a path in this graph that visits every node exactly once? If yes, give the path. \newline
        %\text{graph\_type}: undirected \newline
        \text{thought}: This is  finding a Hamiltonian path.
        ... \newline
        \text{code}:
        \begin{verbatim}
    def hamiltonian_path(graph, path):
    ...
        \end{verbatim}
     % \}
\end{tcolorbox}
% \end{minipage}

\textbf{Documentation.} We crawl the documentation of relevant Python libraries from the Internet, and each entry in the documentation corresponds to a specific application interface (API). Specifically, each entry includes fields such as API name, description, parameters, returns, and examples. 

\textbf{Experience.} We use the system without experience base as the solver, and apply it on the training set to collect experiences. Note that each problem in the training set $\mathcal{D}_\text{train}$ is usually associated with problem type $t\in \mathcal{T}$. Thus we collect candidate experiences $\mathcal{K}_{exp}^t$ for each problem type $t$, and evaluate the utility of each experience on the validation set $\mathcal{D}_\text{valid}$. Only the experience with the highest utility will be kept in each $\mathcal{K}_{exp}^t$. More details including the pseudo codes for building the experience base can be found at Alg.~\ref{alg:problem-solving-experience} in Appendix~\ref{subsec:problem-solving-experience-collection}. 

\subsubsection{Knowledge Retrieval} For effective information retrieval, we build vector databases with LlamaIndex~\cite{llamaindex} for documentation and experiences, respectively. Then the search agent will take the refined question $q_r$ and graph type $q_t$ as query, and find the most relevant experience and API information from the database. Note that the experiences of similar problems contains detailed solutions, and are usually more helpful for downstream agents than documentations. Therefore, we first use similarity matching via LlamaIndex to score the experiences. If the largest similarity score exceeds a predefined threshold $\delta$, the corresponding experience will be directly returned by the search agent without querying related documentation information. Otherwise, the search agent will return the most related entries in documentation base. Finally, either the experience or the documentation information returned by the search agent will be passed to the downstream agents for retrieval-augmented generation (RAG). For simplicity, we denote the operation of the search agent as 
\begin{equation}
    \text{knowledge}_q \gets\operatorname{Search}(q_r,q_t,\mathcal{K},\delta).
\end{equation}

\renewcommand{\algorithmicrequire}{\textbf{Input:}}
\renewcommand{\algorithmicensure}{\textbf{Output:}}

\subsection{Coding and Reasoning Agents} 
Given the outputs of question and search agents, the coding agent will generate and execute Python codes to solve the problems. In practice, the generated codes may encounter compilation or running errors. Following the trial-and-error strategy~\cite{how_to_solve_it, investigating_learning_outcomes} for problem solving, we introduce a retry mechanism that attempts to fix the codes with previous codes and error messages. More formally, the operation of coding agent in the $n$-th trial can be written as:
% {pai ban in Chinese}
\begin{equation}
\begin{split}
        (\text{result}_q^n,\text{code}_q^n,\text{error}_q^n)\leftarrow \operatorname{Coding}(q_r,q_t,q_g,q_f, \\ \text{knowledge}_q, \{\text{code}_q^i\}_{i=1}^{n-1}, \{\text{error}_q^i\}_{i=1}^{n-1}),
\end{split}
\end{equation}
where $\text{result}_q^n,\text{code}_q^n,\text{error}_q^n$ are respectively the execution result, generated code, and error message of the $n$-th trial. The last two parameters are the codes and error messages from previous $n-1$ trials. If the codes runs normally, we define $\text{error}_q^n$ as $\tt None$.

If the coding agent fails to fix the codes when reaching the maximum number of trials, we will employ the reasoning agent to directly answer the question without programming:
\begin{equation}
    \text{result}_q \gets\operatorname{Reasoning}(q_r,q_t,q_g,q_f).
\end{equation}
In this way, the coding and reasoning agents can solve problems collectively, and achieve a collaborative functionality at the system level.

\subsection{Answer Agent} Based on the output format $q_f$, the answer agent will organize the results given by the coding or reasoning agents as final results. Similar to the retry mechanism of coding agent, we also introduce a self-checking mechanism that iteratively refines the results according to format requirement $q_f$. Formally, the operation of answer agent is:
\begin{equation}
        \text{output}_q^n\leftarrow \operatorname{Answer}(q_f,\text{output}_q^{n-1}),
\end{equation}
where $\text{output}_q^n$ is the output in the $n$-th iteration, and $\text{output}_q^0$ is $\text{result}_q$ generated by the coding or reasoning agents.

% \begin{minipage}[t]{0.46\textwidth}
\begin{tcolorbox}[colback=gray!10, colframe=black, rounded corners, boxrule=1.5pt, fontupper=\normalsize, left=2mm, right=2mm, top=1mm, bottom=1mm]
    $\textbf{Input}$ \newline
    $\text{result}$: 0, 1, 6, 3, 5, 4, 2. \newline
    $\text{output\_format}$: Yes. The path can be: 1, 4, 8. \newline
    $\textbf{Answer}$ \newline
    $\text{result}$: Yes. The path can be: 0, 1, 6, 3, 5, 4, 2.
    
\end{tcolorbox}
% \end{minipage}

\subsection{Summary}
We formalize the inference pipeline of \modelname in Alg.~\ref{alg:pipeline}. Most agent functions (including Question, Coding, Reasoning and Answer) in Alg.~\ref{alg:pipeline} are all implemented by an LLM with different prompt templates. The Search function is implemented based on LlamaIndex~\cite{llamaindex}. We will provide more implementation details in Appendix~\ref{sec:methods-details}.

\begin{algorithm}[t]
\caption{Inference Pipeline of \modelname}
\label{alg:pipeline}
\begin{algorithmic}[1]
\REQUIRE Problem described by question $q$, knowledge base $\mathcal{K}$, threshold $\delta$ for determining whether experience or documentation is retrieved, maximum number of retries $N_{retry}$, number of self-checking iterations $N_{check}$.
\ENSURE Final answer of the problem $\text{output}_q$.
\STATE $(q_r,q_t,q_g,q_f)\leftarrow \operatorname{Question}(q)$
\STATE $\text{knowledge}_q \gets\operatorname{Search}(q_r,q_t,\mathcal{K},\delta)$
\STATE $n=0$
\WHILE{$\text{error}_q^n \neq \tt{None}$ and $n < N_{retry}$}
\STATE ++n

\STATE $(\text{result}_q^n, \text{code}_q^n, \text{error}_q^n) \gets \operatorname{Coding}(q_r, q_t, q_g, q_f, $ $ \text{knowledge}_q,
    \{\text{code}_q^i\}_{i=1}^{n-1}, \{\text{error}_q^i\}_{i=1}^{n-1})$
\STATE $\text{result}_q \gets\text{result}_q^n$
\ENDWHILE

\IF{$\text{error}_q^n \neq \tt{None}$}
    \STATE $\text{result}_q \gets\operatorname{Reasoning}(q_r,q_t,q_g,q_f)$
\ENDIF

\STATE $\text{output}_q^0 \gets \text{result}_q$
\FOR{$n = 1, 2, \dots, N_{check}$}
        \STATE $\text{output}_q^n \gets \operatorname{Answer}(q_f, \text{output}_q^{n-1})$
\ENDFOR
\end{algorithmic}
\vspace{-0.2em}
\end{algorithm}

\section{Experiments}
We conduct extensive experiments on six graph reasoning benchmarks to validate the effectiveness of  \modelname.

\subsection{Experimental Setup}
\label{subsec:experiment_setup}

\subsubsection{Datasets}

To evaluate the effectiveness of our system, we consider six benchmarks for LLM-based graph reasoning: Talk like a Graph~\cite{talk_like_a_graph} primarily studies some fundamental problems in graph reasoning and explores the cognitive abilities of LLMs regarding graphs. LLM4DyG~\cite{LLM4DyG} focuses on basic problems related to dynamic graphs, while GraphWiz~\cite{graphwiz}, NLGraph~\cite{nlgraph}, and GraphInstruct~\cite{graphinstruct} are used to evaluate the understanding and mastery of algorithms in fundamental graph theory. GNN-AutoGL is a benchmark we constructed in this paper to study the ability of LLMs in deploying GNNs with AutoGL~\cite{AutoGL}. Dataset statistics and details can be found at Appendix~\ref{sec:dataset-statistics} and ~\ref{sec:gnn_autogl_benchmark}.

\subsubsection{Metrics}

For benchmarks with definite answers~\cite{talk_like_a_graph, nlgraph, LLM4DyG, graphwiz, graphinstruct}, we use accuracy as the evaluation metric. For GNN-AutoGL, since the task requires a complex pipeline (including identifying proper APIs, filling correct hyper-parameters, successfully training and evaluating the corresponding GNN model on a specified dataset, we evaluate whether the generated codes can be executed correctly and whether the key hyper-parameters (\textit{e.g.,} dataset name, model name, hidden dimension, dropout) are correct as the evaluation criteria. More details are shown in Appendix~\ref{sec:metrics-details}.

\subsubsection{Settings}

We choose GPT-4o-mini~\cite{gpt_4o_mini} and Qwen2.5-Coder-7B~\cite{qwen25codertechnicalreport} as the base LLM for the agents of \modelname. For Python libraries, we use NetworkX version 3.3~\cite{NetworkX} and AutoGL version 0.4.0~\cite{AutoGL}, together with Python version 3.10.14. The hyper-parameters are set as follows: the number of candidate experiences $N_{exp} = 10$, similarity matching threshold for using experience $\delta= 0.85$. The maximum number of retries for the coding agent is $3$, and the number of self-checking iterations for the answer agent is $3$.

\subsubsection{Baselines} 
As the SOTA methods for each benchmark are quite diverse and some of which are designed for a specific benchmark, we directly compare with the \textit{previously reported SOTA} performance for each benchmark: for NLGraph, we select GUNDAM~\cite{GUNDAM} as the SOTA; for Talk like a Graph, we select the fine-tuned transformers~\cite{transformer_reasoning} as the SOTA; for LLM4DyG, GraphWiz, and GraphInstruct~\cite{LLM4DyG, graphwiz, graphinstruct}, we select the best model in their benchmark papers. In addition, we adopt AutoGPT~\cite{autogpt} and Open Interpreter~\cite{openinterpreter} as single-agent baselines, while employing AutoGen~\cite{autogen} and ChatDev~\cite{ChatDev} as multi-agent baselines for comparison.

\subsection{Main Results}
\label{subsec:main_results}

\begin{table*}[!ht]
    \vspace{-1em}
  \caption{Performance comparison between \modelname and baselines in terms of accuracy (\%).}
  \label{tab:main-experiments}
  % \vspace{-1em}
  \centering
  \rowcolors{2}{white}{gray!15}
  \resizebox{1.0\textwidth}{!}{%
  \begin{tabular}{llcccccc|c}
    \toprule
    Base LLM & 
    \hspace{1em} Methods & NLGraph & \makecell{Talk like a Graph} & GraphInstruct & GraphWiz &  LLM4DyG & GNN-AutoGL & Average Rank \\
    \midrule 
    \multirow{-2}{*}{\cellcolor{white}} - & \hspace{1em} Previous SOTA & 79.7 & \underline{90.2} & 43.5 & 65.0 & 53.7 & - & 5.5 \\
    \midrule
    \cellcolor{white} & \hspace{1em} Base Model & 51.4 & 61.3 & 53.7 & 47.6 & 58.0 & 3.3 & 7.5 \\
    \cellcolor{white} & \hspace{1em} AutoGPT & 65.8 & 65.2 & 58.5 & 50.2 & 61.6 & 10.3 & 5.7 \\
    \cellcolor{white} & \hspace{1em} Open Interpreter & 68.2 & 85.6 & 54.2 & 53.9 & 55.5 & 8.8 & 5.5 \\
    \cellcolor{white} & \hspace{1em} AutoGen & 59.7 & 70.2 & 57.7 & 62.8 & 47.3 & 28.1 & 5.8 \\
    \cellcolor{white} & \hspace{1em} ChatDev & 79.6 & 72.5 & 70.6 & 67.8 & 77.8 & 34.5 & 3.5 \\
    % \rowcolor{rowgray2}
    \multirow{-5}{*}{\cellcolor{white}\makecell[l]{GPT-4o-mini}} & \hspace{1em} \textbf{GraphTeam} & \textbf{97.8} & \textbf{94.5} & \textbf{88.1} & \textbf{88.6} & \textbf{94.8} & \textbf{48.4} & \textbf{1.0} \\
    \midrule
    
    % 知识检索类别
    % \rowcolor{rowgray1}
    \cellcolor{white} & \hspace{1em} Base Model & 37.3 & 63.1 & 34.5 & 51.8 & 38.2 & 1.9 & 8.3 \\
    % \rowcolor{rowgray2}
    \multirow{-2}{*}{\cellcolor{white}\makecell[l]{Qwen2.5-Coder-7B}} & \hspace{1em} \textbf{GraphTeam} & \underline{93.0} & 85.6 & \underline{84.5} & \underline{78.6} & \underline{87.4} & \underline{36.7} & \underline{2.2} \\
    \bottomrule
  \end{tabular}
  }
   % \vspace{-0em}
\end{table*}

As shown in Table~\ref{tab:main-experiments}, \modelname achieves significant and consistent improvements over baseline methods on all six benchmarks. Compared with the best baseline, \modelname (GPT-4o-mini) has 15.3\% accuracy gains on average. Even when equipped with open-source model, \modelname (Qwen2.5-Coder-7B) still has 7.5\% average accuracy improvements over the best baseline based on GPTs. These results demonstrate that \modelname can achieve promising performance with both open-source and closed-source LLMs. Among the baselines, LLM-based agents (\textit{i.e.,} AutoGPT, Open Interpreter, AutoGen and ChatDev) consistently achieve better performance than the corresponding base LLM (\textit{i.e.,} GPT-4o-mini). For example, the single-agent baseline AutoGPT has 6.1\% average accuracy improvements over the base LLM. Moreover, the multi-agent baseline ChatDev achieves even better results, with 21.3\% average accuracy gains over the base LLM. These observations support our motivation of employing LLM-based agents for graph reasoning.

\subsection{Ablation Study}
\label{subsec:ablation_study}

To evaluate the contribution of each component, we conduct a systematic ablation study of \modelname using GPT-4o-mini. We assess the impact of \textit{Input-Output Normalization} by removing the question or answer agent. We examine the influence of \textit{Knowledge Retrieval} components in the search agent, specifically documentation and experience. Additionally, the effectiveness of \textit{Problem Solving} designs are evaluated by eliminating the coding or reasoning agent, as well as the retry mechanism from the coding agent. This comprehensive ablation enables us to quantify each component's significance to the system performance. The results are presented in Table~\ref{tab:ablation-study-experiments}.

\begin{table*}[!ht]
    \vspace{-1em}
  \caption{\small Performance comparison between GraphTeam and ablation variants in terms of accuracy (\%).}
  \label{tab:ablation-study-experiments}
  % \vspace{-1em}
  \centering
  \rowcolors{2}{white}{gray!15}
  \resizebox{1.0\textwidth}{!}{%
  \begin{tabular}{llcccccc}
    \toprule
    Category & Methods & NLGraph & \makecell{Talk like a Graph} & GraphInstruct & GraphWiz & LLM4DyG & GNN-AutoGL \\
    \midrule
      
    \multirow{-2}{*}{\cellcolor{white}}& {\textbf{\modelname (GPT-4o-mini)}} & \textbf{97.8} & \textbf{94.5} & \textbf{88.1} & \textbf{88.6} & \textbf{94.8} & \textbf{48.4} \\
    \midrule
    \cellcolor{white} & \hspace{1em}- question agent & 79.2 & 82.1 & 85.9 & 62.8 & 58.6 & 14.8 \\
    % \rowcolor{rowgray2}
    \multirow{-2}{*}{\cellcolor{white}\makecell[l]{Input-Output\\Normalization}} & \hspace{1em}- answer agent & 96.7 & \underline{93.9} & 85.3 & \underline{88.0} & \underline{94.2} & 45.8 \\
    \midrule
    
    % 知识检索类别
    % \rowcolor{rowgray1}
    \cellcolor{white} & \hspace{1em}- documentation & 96.4 & 93.8 & \underline{86.1} & \underline{88.0} & 94.0 & \underline{46.9} \\
    % \rowcolor{rowgray2}
    \multirow{-2}{*}{\cellcolor{white}\makecell[l]{Knowledge\\ Retrieval}} & \hspace{1em}- experience & 54.7 & 92.0 & 76.6 & 82.8 & 60.3 & 4.8 \\
    % 输入输出归一化类别
    % \rowcolor{rowgray1}
    % \multirow{-1}{*}
    % \midrule
    % 问题解决类别
    % \rowcolor{rowgray2}
    \midrule
    \cellcolor{white}& \hspace{1em}- coding agent & 39.3 & 60.7 & 61.1 & 46.4 & 59.3 & - \\
    % \rowcolor{rowgray1}
    \cellcolor{white}& \hspace{1em}- retry & 87.7 & 93.5 & 85.7 & 83.5 & 88.5 & 44.3 \\
    % \rowcolor{rowgray2}
    \multirow{-3}{*}{\cellcolor{white}\makecell[l]{Problem\\Solving}} & \hspace{1em}- reasoning agent & \underline{96.9} & \underline{93.9} & 85.5 & 54.2 & 93.7 & 46.4 \\
    \bottomrule
  \end{tabular}
  }
   % \vspace{-0em}
\end{table*}

\textbf{Input-Output Normalization}: For most benchmark problems, the \textbf{question agent} significantly contributes (25.3\%) by breaking the problem into individual components on average. For a few benchmark such as GraphInstruct~\cite{graphinstruct}, where the structure of the question is already clear and straightforward, the impact of the question agent is less noticeable (less than 2.2\%). The \textbf{answer agent}'s correction of the format is also an essential part of correctly answering the questions (0.6-2.8\%).

\textbf{External Knowledge Retrieval}: The \textbf{search agent} plays a significant role. When solving graph reasoning problems, providing LLM with related problem-solving experiences or API documentation can help the model solve the problem more effectively. The impact of \textbf{experience} is greater than that of documentation, indicating that LLMs solve problems in a way similar to humans with an average improvement of 23.5\%. They can derive better solutions from prior experiences, effectively addressing similar problems. When relevant experiences are not available, \textbf{documentation} can also help enhance the model's programming capabilities in solving corresponding problems (0.6\% to 2.0\%).% \textbf{Documentation} can enhance the model's programming capabilities. 

\textbf{Problem Solving}: The \textbf{coding agent} plays a crucial role. Without programming and relying solely on reasoning, the system's performance would significantly decrease, indicating that programming is an effective approach for solving graph reasoning problems (27.0\% to 58.5\%). The \textbf{retry} mechanism also brings a positive effect (1.0\% to 10.1\%) by debugging programs. The \textbf{reasoning agent} does not have a significant impact on problems where programming achieves good results, but it plays an important role in problems where programming is insufficient and reasoning is required (0.9\% to 34.4\%).

The contribution of each component of the system varies across different benchmarks. Since the core problem-solving component of this system is the coding agent, programming to solve problems is a crucial and effective approach for graph reasoning. For GNNs, in particular, coding is an indispensable component. Therefore, removing the coding agent leads to a significant drop in performance for all benchmarks (40.9\% on average). Some benchmarks do not have strict requirements for the answer format (Talk like a graph, GraphWiz, and LLM4DyG), thus removing the answer agent does not significantly impact the overall system performance. Since questions in Talk like a Graph are relatively simple, most problems can be correctly solved through coding without the need for the reasoning agent. Hence this benchmark has minimal need for the reasoning agent. In summary, the full model always perform the best, validating the necessity of each design.

\subsection{Analysis on Different Groupings}
\label{subsec:analysis_on_different_goupings}

\begin{figure*}[htbp]
    \centering
    \includegraphics[width=1.0\linewidth]{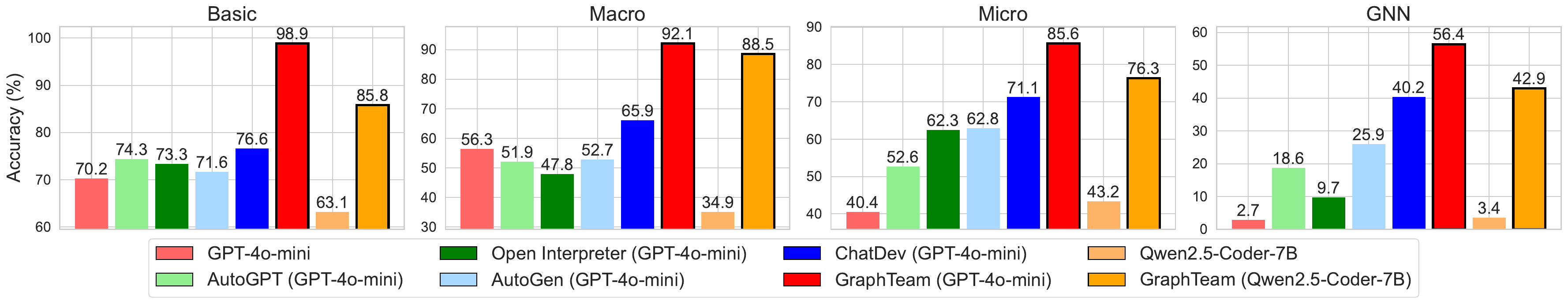}
    \vspace{-1.5em}
    \caption{Performance with respect to different task categories.}
    \label{fig:task_catogory}

\end{figure*}

\textbf{Group by Task Categories.}
We classify the questions into four categories based on the nature of the task: (1) basic: fundamental graph understanding, (2) macro-level: reasoning over entire graph structures, (3) micro-level: local properties of nodes and edges, and (4) GNN: graph neural networks. As shown in Figure~\ref{fig:task_catogory}, \modelname (GPT-4o-mini) achieves over {90\%} accuracies on the first two categories, and {85.6\%} on micro-level tasks, ranking the first over all four groups. While \modelname (Qwen2.5-Coder-7B) always ranks the second, achieving over {76\%} on the first three categories. These results indicate that our framework can effectively handle common graph reasoning tasks. For most categories (basic, micro-level, and GNN), we observe consistent improvements when utilizing agent frameworks. But on macro-level tasks, we observe that AutoGPT, Open Interpreter, and AutoGen perform worse than using GPT-4o-mini alone. We find that this degradation is caused by misunderstanding question requirements during long reasoning processes. Thus, employing the question agent to explicitly identify key information is necessary for graph reasoning. Besides, all systems struggle with GNN-related tasks, suggesting a need for improvement in future work. 

\begin{figure*}[htbp]
    \centering
    \includegraphics[width=1.0\linewidth]{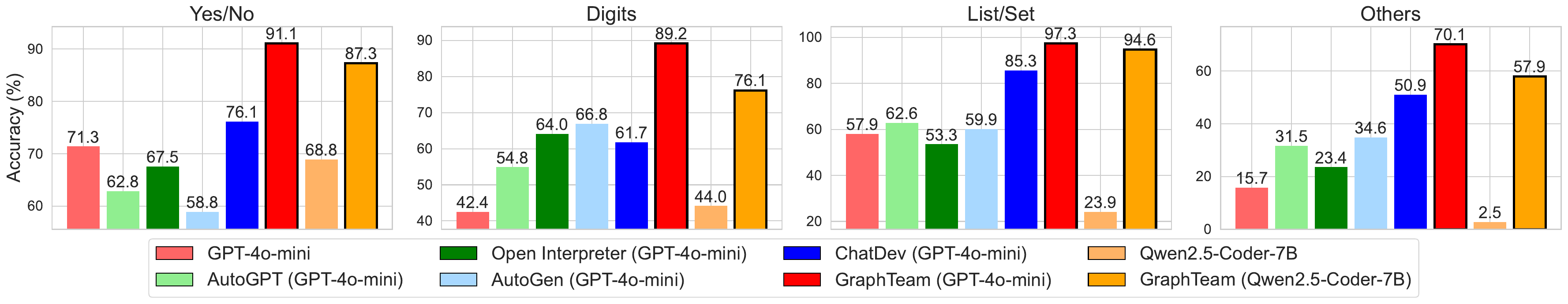}
    \vspace{-1.5em}
    \caption{Performance with respect to different output formats.}
\label{fig:output_formats}
\end{figure*}

\textbf{Group by Output Types.} We categorize problems according to their required data formats: true/false questions (yes/no), numerical calculations (digits), analytical questions (list/set), and other types, with results presented in Figure~\ref{fig:output_formats}. Similarly, \modelname with GPT-4o-mini and Qwen2.5-Coder-7B always rank the first and second across all four output formats. In particular, \modelname (GPT-4o-mini) reaches or approaches 90.0\% for the first three categories, showing a strong generalization ability. The base LLMs perform good for true/false questions, but are inferior to agent-based methods for problems requiring more complex output formats.

\begin{figure}[!ht]
    \centering
    \includegraphics[width=1.0\linewidth]{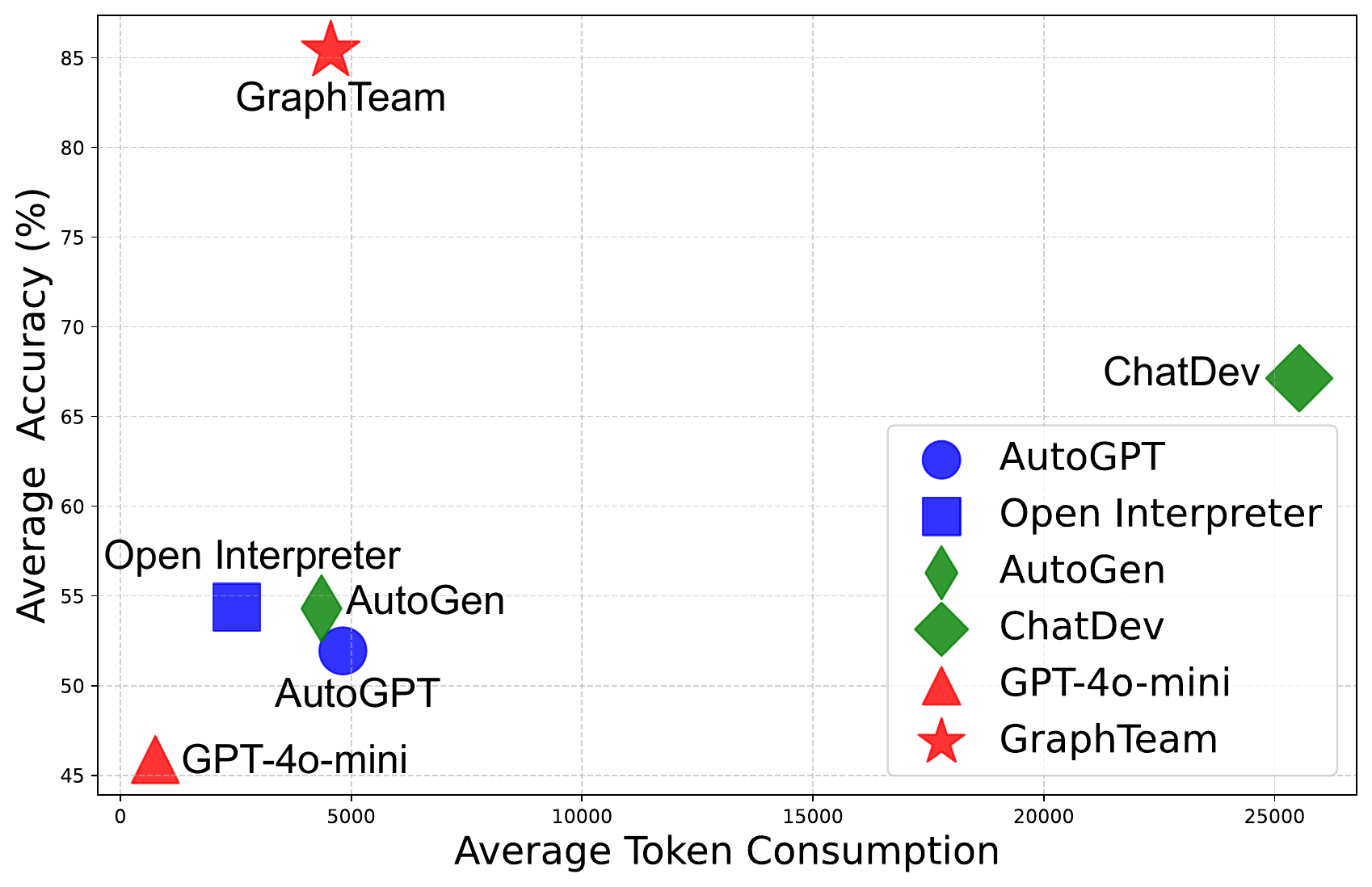}
    \vspace{-1.2em}
    \caption{Average accuracies (\%) and token consumptions of different agent-based methods. While maintaining comparable token consumption with typical agent frameworks, \modelname substantially improves the average accuracy for graph reasoning.}
    \label{fig:token_and_accuracy}
\end{figure}

\subsection{Efficiency Analysis}
\label{subsec:efficiency_analysis}

To verify whether our proposed \modelname is cost effective, we present average accuracy and token consumption for all agent-based methods based on GPT-4o-mini in Figure~\ref{fig:token_and_accuracy}. \modelname achieves an average accuracy of 85.4\% with relatively low token consumption. The token cost of \modelname is competitive with single-agent baselines, and is significantly lower than ChatDev. Though ChatDev demonstrates notable accuracy advantages over other agent frameworks, it consumes approximately 5.6 times more tokens than \modelname. This indicates that our framework design is better optimized for graph reasoning tasks, achieving promising performance with acceptable resource consumption.

\subsection{Hyper-parameter Analysis}
\label{subsec:hyper-parameter_analysis}

\begin{figure*}[!ht]
    % \small
    \centering
    \vspace{-1em}
    \begin{subfigure}
        \centering
        \label{fig:experience}
        \includegraphics[width=0.23\linewidth]{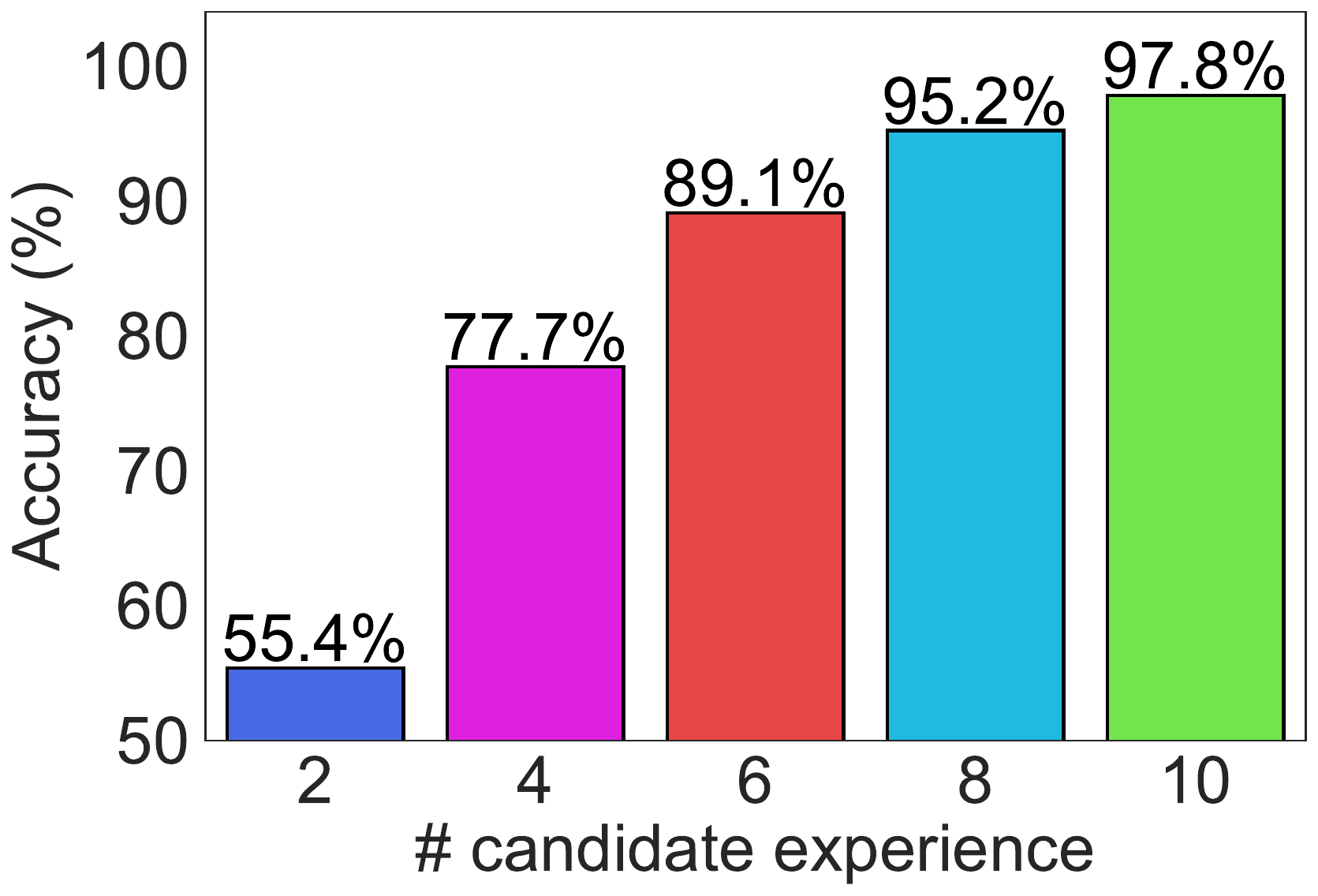}
    \end{subfigure}
    \hfill
    \begin{subfigure}
        \centering
        \label{fig:threshold}
        \includegraphics[width=0.23\linewidth]{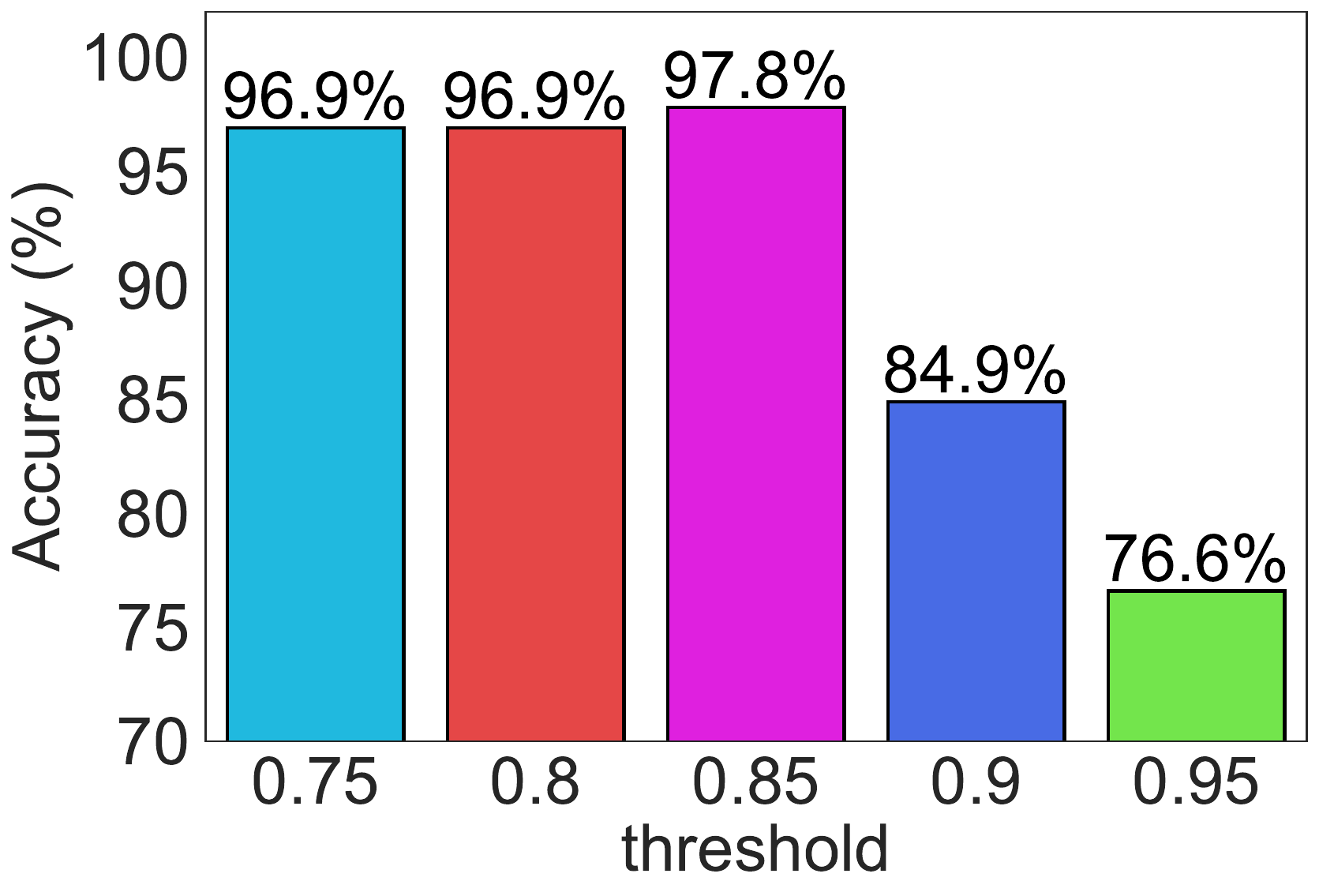}
    \end{subfigure}
    \hfill
    \begin{subfigure}
        \centering
        \label{tab:coding_agent_retry}
        \includegraphics[width=0.23\linewidth]{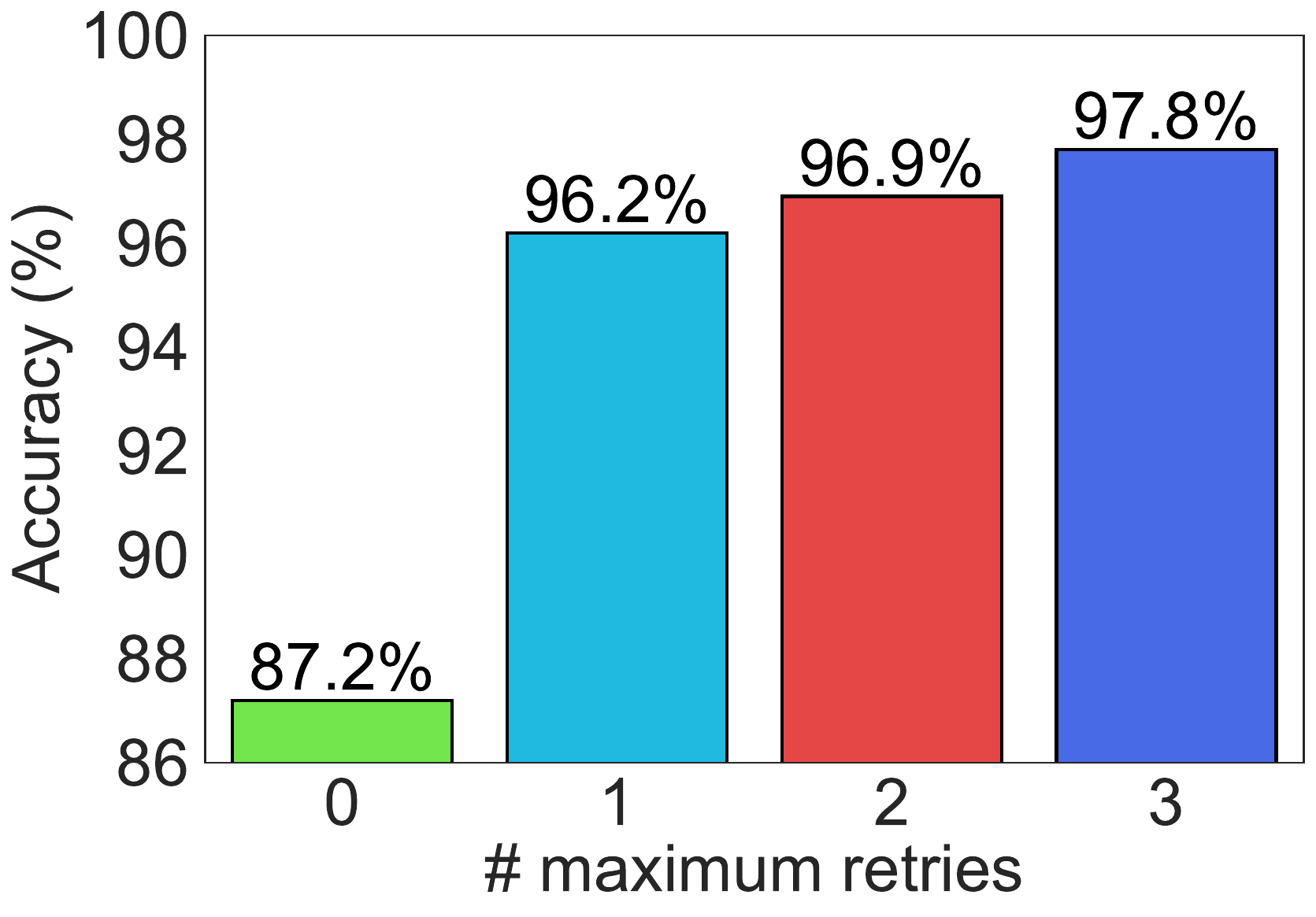}
    \end{subfigure}
    \hfill
    \begin{subfigure}
        \centering
        \label{tab:answer_agent_self_check}
        \includegraphics[width=0.23\linewidth]{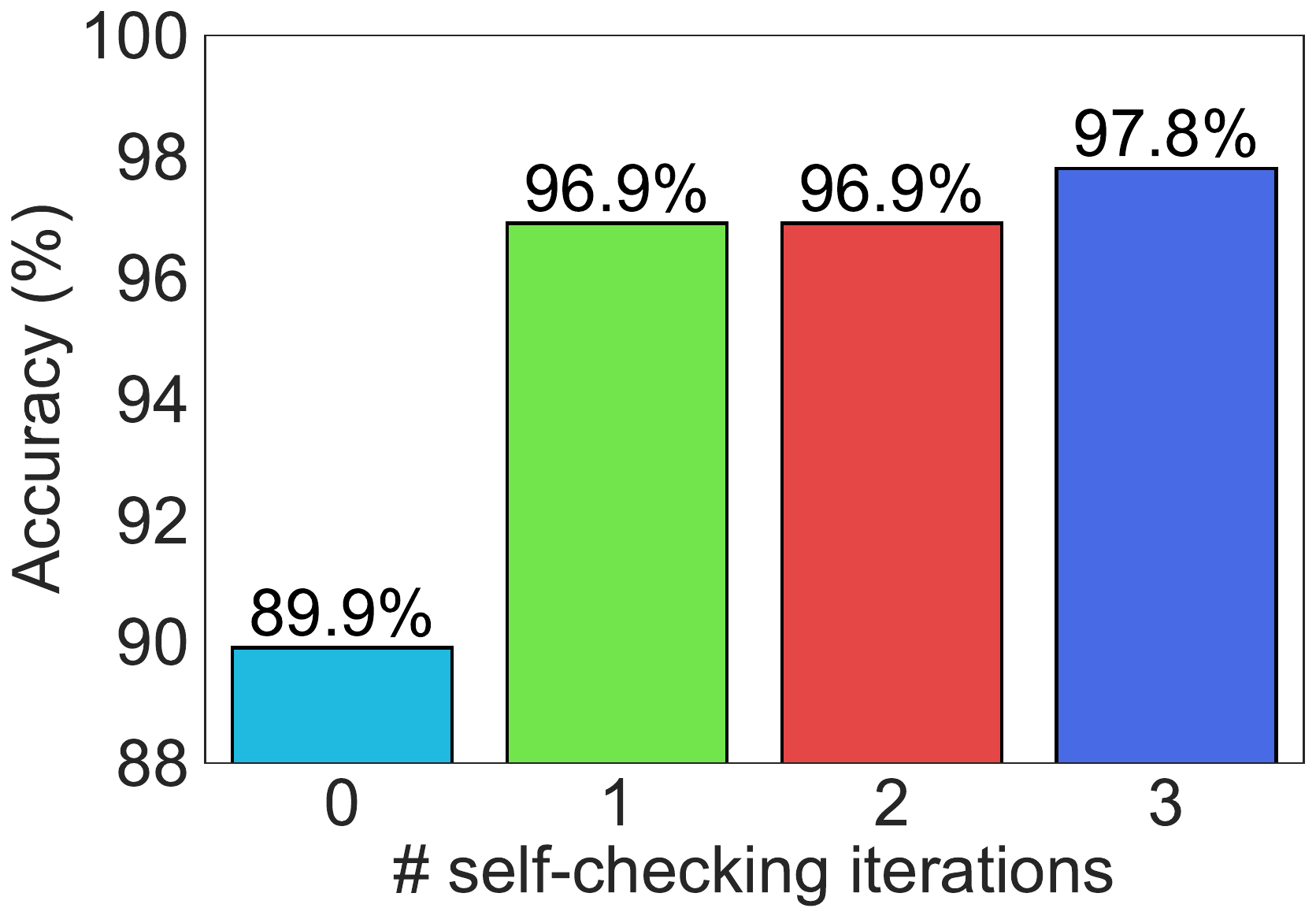}
    \end{subfigure}
    \vspace{-1.5em}
    \caption{Analysis of key hyper-parameters in the proposed \modelname.}
    \label{fig:hyper-parameter}
    \vspace{-0.5em}
\end{figure*}

As shown in Figure~\ref{fig:hyper-parameter}, we conduct experiments to study the influence of key hyper-parameters using the NLGraph~\cite{nlgraph} benchmark. For the number of candidate experiences $N_{exp}$ in knowledge base construction, our experiments show that larger $N_{exp}$ values correlate with better system performance, as they provide more candidate experiences for problem-solving assistance. The similarity matching threshold $\delta$ between experiences and documents achieves optimal problem-solving performance at $0.85$, enabling correct matching of similar experiences when available and document retrieval when experiences are absent. Increasing the maximum retries of the coding agent and the number of self-checking iterations of the answer agent improves system performance, but with diminishing returns as these values increase, eventually plateauing. These findings provide valuable insights for optimizing the system's configuration and balancing performance gains against computational costs.

\section{Conclusion}
% \input{GraphTeam/conclusion}
% \section{Conclusion}
In this work, we draw inspiration from human problem-solving processes, and design a multi-agent system named \modelname for graph reasoning. \modelname consists of five agents across three key modules, namely input-output normalization, external knowledge retrieval, and problem solving. The agents can cooperate with each other for solving complex graph reasoning problems. Extensive experiments show that \modelname achieves state-of-the-art results with impressive accuracy improvements. 

As the first work to address LLM-based graph reasoning problems using a multi-agent system, our workflow prioritizes simplicity, with most agents operating sequentially. For future work, we will consider to explore more sophisticated communications between agents, and automatically optimize the system pipeline to accommodate more tasks. It is also possible to fine-tune the LLMs in \modelname for better performance. 

\medskip

\bibliographystyle{plainnat}
\bibliography{reference}

%%%%%%%%%%%%%%%%%%%%%%%%%%%%%%%%%%%%%%%%%%%%%%%%%%%%%%%%%%%%

\appendix

% \input{GraphTeam/app_method}

% \newpage

% \;

\newpage

\section{Method Details}
\label{sec:methods-details}

\subsection{Problem-Solving Experience Collection}
\label{subsec:problem-solving-experience-collection}

The pseudo code for building the experience base is shown in Alg.~\ref{alg:problem-solving-experience}. Each problem in the training or validation datasets is represented as a triple $(q,a,t)$ of question, answer and type. The solver is formalized as function $\operatorname{Solver}(\operatorname{Question}(q),e)$ with the output of question agent and an optional experience entry $e$ as inputs, and can be seen as a simplified version of \modelname without the entire experience base. Besides generated answer, the outputs of the solver also include meta information such as codes and thinking processes. If the solver can correctly answer a problem, we will add a triple $(q,a,\text{meta\_info})$ of question, answer and meta information into the candidate experiences $\mathcal{K}_{exp}^t$. Then for each experience $e\in\mathcal{K}_{exp}^t$, we evaluate its utility by solving the problems of the same type $t$ in the validation set $\mathcal{D}_\text{valid}$ with experience $e$ as augmentation. Then the experience base $\mathcal{K}_{exp}$ collects the experience with the highest utility in each $\mathcal{K}_{exp}^t$.

\begin{algorithm}[htbp]
% \vspace{-0.5em}
% \small
\caption{Problem-Solving Experience Collection}
\label{alg:problem-solving-experience}
% \vspace{-0.2em}
\begin{algorithmic}[1]
\REQUIRE Training dataset $\mathcal{D}_\text{train}$, validation dataset $\mathcal{D}_\text{valid}$, candidate experience size $N_{exp}$ for each problem type $t\in\mathcal{T}$.
\ENSURE Experience knowledge base $\mathcal{K}_{exp}$.

\FOR{each problem $(q,a,t)$ from $\mathcal{D}_\text{train}$}
        \IF {$|\mathcal{K}_{exp}^t|$ $<$ $N_{exp}$}
            \STATE $(\text{answer},\text{meta\_info}) \gets \text{Solver}(\operatorname{Question}(q))$
            \IF{answer equals to ground truth $a$}
                \STATE $\mathcal{K}_{exp}^t \gets \mathcal{K}_{exp}^t \cup \{(q,a, \text{meta\_info})\}$
            \ENDIF
        \ENDIF
\ENDFOR

\FOR{each problem $(q,a,t)$ from $\mathcal{D}_\text{valid}$}
\FOR{$e \in \mathcal{K}_{exp}^t$}
\STATE $(\text{answer},\text{meta\_info}) \gets \text{Solver}(\operatorname{Question}(q),e)$           
            \IF{answer equals to ground truth $a$}
                \STATE $\text{Utility}_e \gets \text{Utility}_e+1$
            \ENDIF
\ENDFOR
\ENDFOR

\FOR{$t \in \mathcal{T}$}
        \STATE $\mathcal{K}_{exp}^t \gets \text{argmax}_{e\in \mathcal{K}_{exp}^t}  \text{Utility}_e$
\ENDFOR

\STATE $\mathcal{K}_{exp} \gets \{ \mathcal{K}_{exp}^t\}_{t \in \mathcal{T} }$
%\Comment{Define $\mathcal{E}$ as the collection of all $\mathcal{E}_t$}

\end{algorithmic}
% \vspace{-0.2em}
\end{algorithm}

\subsection{Prompt Templates}

Below are the prompts of the five agents used in \modelname. 

\subsubsection{Question Agent}

% \begin{minipage}[t]{0.46\textwidth}
\begin{tcolorbox}[colback=gray!10, colframe=black, rounded corners, boxrule=1.5pt, fontupper=\normalsize, left=2mm, right=2mm, top=1mm, bottom=1mm]
    % \textbf{Requirement Analyst} \newline
    Here is the task: \{original question\}. \newline
    As a requirement analyst, you need to extract key information from the task, specifically focusing on the following elements: \newline
    Reformatted\_Problem (str): This is the specific problem that needs to be solved. \newline
    Graph\_Type (str): This refers to the type of graph mentioned in the task, such as directed, undirected, weighted, dynamic, etc. If the task does not specify a graph type, it should be set to undirected. \newline
    
\end{tcolorbox}
% \end{minipage}

\begin{tcolorbox}[colback=gray!10, colframe=black, rounded corners, boxrule=1.5pt, fontupper=\normalsize, left=2mm, right=2mm, top=1mm, bottom=1mm]
    Input\_Data (str): This refers to the graph data or graph file required by the task. You must extract \textbf{all relevant information} about the graph data, including but not limited to graph representation (adjacency list, adjacency matrix, edge list, etc.), node and edge details, and any other relevant information. \newline
    Output\_Format (str): This is the format of the output answer mentioned in the task. If the task does not specify a format, it should be set to None. \newline
    Note that we must \textbf{ONLY} discuss these key elements and do not discuss anything else! Please provide the extracted information in JSON format. Your answer must be in \textbf{JSON} format. 
    
\end{tcolorbox}

\subsubsection{Search Agent}

% \begin{minipage}[t]{0.46\textwidth}
\begin{tcolorbox}[colback=gray!10, colframe=black, rounded corners, boxrule=1.5pt, fontupper=\normalsize, left=2mm, right=2mm, top=1mm, bottom=1mm]
    You are tasked as a {research assistant} to retrieve relevant information for the task: \{reformatted\_question, graph\_type\}. \newline
    Use the available knowledge base and memory to find the most accurate and relevant information related to the task. \newline
    If the query is too ambiguous, try to clarify it using previous context. \newline
    Please provide the results in a structured format.
    
\end{tcolorbox}
% \end{minipage}

\subsubsection{Coding Agent}

% \begin{minipage}[t]{0.46\textwidth}
\begin{tcolorbox}[colback=gray!10, colframe=black, rounded corners, boxrule=1.5pt, fontupper=\normalsize, left=2mm, right=2mm, top=1mm, bottom=1mm]
    % \textbf{Graph Learning Specialist} \newline
    Based on the available search result: \{konwledge of the question\}, your goal is to write Python codes that solve the specified graph learning problem using libraries like NetworkX. \newline
    If there is no search result, generate the code based on your expertise. \newline
    ATTENTION: This is the actual problem that needs to be solved, based on the provided information. Please disregard the example above while solving this real task.\newline 
    Reformatted\_Problem: {reformatted\_problem}, \newline
    Graph\_Type: {graph\_type} \newline 
    Input\_Data: {input\_data}, \newline 
    Output\_Format: {output\_format}. \newline
    The output code must adhere to the following format:\begin{verbatim}
    ```python
    ......
    ```
    \end{verbatim}
    to ensure easy extraction. \newline
    If the code execution failed. Please analyze the error message and the code provided to identify the issuescausing the failure in code execution. Based on your analysis, revise the code and provide a corrected version that should execute successfully. \newline
    Here is the error message: {error\_message} \newline
    The revised code must be formatted as follows for easy extraction: 
    \begin{verbatim}
    ```python
    ......
    ```
    \end{verbatim}
    
\end{tcolorbox}
% \end{minipage}

\subsubsection{Reasoning Agent}

% \begin{minipage}[t]{0.46\textwidth}
\begin{tcolorbox}[colback=gray!10, colframe=black, rounded corners, boxrule=1.5pt, fontupper=\normalsize, left=2mm, right=2mm, top=1mm, bottom=1mm]
    % \textbf{Graph Learning Expert} \newline
    As a {graph learning expert}, your task is to analyze and solve a graph reasoning problem: \{refomatted\_question\}. \newline
    Despite any challenges that previous attempts may have encountered, your goal is to derive a solution leveraging your expertise in graph theory and machine learning. \newline
    Here is the task based on the extracted information: Input\_Data: {input\_data}, Reformatted\_Problem: {reformatted\_problem}, Output\_Format: {output\_format}. \newline
    Provide a clear, concise answer and present your solution in the following format for easy extraction:
    
\end{tcolorbox}
% \end{minipage}

% \newpage

\subsubsection{Answer Agent}

% \begin{minipage}[t]{0.46\textwidth}
\begin{tcolorbox}[colback=gray!10, colframe=black, rounded corners, boxrule=1.5pt, fontupper=\normalsize, left=2mm, right=2mm, top=1mm, bottom=1mm]
    % \textbf{Output Format Specialist} \newline
    As an {assistant\_role}, your role is to ensure that the generated output strictly adheres to the specified format requirements. \newline
    You will be provided with both the required output format (example format) and the actual output. Your task is to compare the two and determine if the actual output matches the format requirements. If the output does not meet the format, you must adjust the format accordingly while \textbf{preserving every part of the original content, meaning, and intent without deleting or changing any sections}. \newline
    Important: The required format provided is an \textbf{example} that illustrates the expected structure, not the actual content or final answer. Do \textbf{not} alter the original meaning, sections, sentences, or intent of the actual output. Your task is only to adjust the format if necessary. \newline
    When adjusting the format, ensure that \textbf{no parts are removed, shortened, or reworded in a way that changes the original meaning or omits content}. The adjustment should focus on structural changes such as spacing, indentation, or organization. \newline
    Once the adjustment is made, compare the adjusted output with the original to validate that \textbf{all sections, sentences, and words are fully preserved} in the final output. \newline
    Here are the provided details: \newline
    - \textbf{Required format (example)}: {output\_format} \newline
    - \textbf{Actual output} (content to be reviewed): {output} \newline
    Carefully review both and ensure that the actual output conforms to the structure of the example format, while \textbf{maintaining the original content, meaning, sections, and intent}. \newline
    Note: Only adjust the \textbf{format} if necessary. Do \textbf{not} change the actual content, meaning, or intent of the output. If the format is already correct, no adjustment is needed. \newline
    Your response should be in JSON format with: \newline
    - \textbf{need\_adjustment} (True/False): Indicating whether an adjustment is required. \newline
    - \textbf{output} (str): If no adjustment is needed, this should match the original output exactly. If an adjustment is needed, this should contain the adjusted output, with only format changes applied, without altering the content, meaning, sections, or intent.
    
\end{tcolorbox}

\section{Details about Datasets and Metrics}

\subsection{Dataset Statistics}
\label{sec:dataset-statistics}

We list the statistics of the six benchmarks in Table~\ref{tab:dataset_statistics}, including task categories, output formats, maximum graph sizes, and the number of samples in experience and test sets.

\begin{table*}[htbp]
\centering
\caption{Statistics of six graph reasoning benchmarks.}
\resizebox{\textwidth}{!}{%
\begin{tabular}{cccccc} 
\toprule
   & \makecell[c]{Task Category} & \makecell[c]{Output Format} & \makecell[c]{Graph Size} & \makecell[c]{Total Experiences} & \makecell[c]{Test Set Size} \\ 
\midrule

NLGraph & \makecell[c]{Macro-level, Micro-level, GNN} & \makecell[c]{yes/no, digits, list/set, others} & \makecell[c]{$\sim 10^1$} & \makecell[c]{80} & \makecell[c]{1,000} \\ 

\makecell[c]{Talk like a Graph} & \makecell[c]{Basic, Macro-level} & \makecell[c]{yes/no, digits, list/set} & \makecell[c]{$\sim 10^1$} & \makecell[c]{80} & \makecell[c]{8,000} \\ 

GraphInstruct & \makecell[c]{Basic, Macro-level, Micro-level} & \makecell[c]{yes/no, digits, list/set} & \makecell[c]{$\sim 10^1$} & \makecell[c]{170} & \makecell[c]{5,100} \\ 

GraphWiz & \makecell[c]{Macro-level, Micro-level} & \makecell[c]{yes/no, digits, list/set} & \makecell[c]{$\sim 10^2$} & \makecell[c]{90} & \makecell[c]{3,600} \\ 

LLM4DyG & \makecell[c]{Basic, Micro-level} & \makecell[c]{yes/no, digits, list/set} & \makecell[c]{$\sim 10^1$} & \makecell[c]{90} & \makecell[c]{2,700} \\ 

GNN-AutoGL & \makecell[c]{GNN} & \makecell[c]{others} & \makecell[c]{$\sim 10^3$} & \makecell[c]{10} & \makecell[c]{200} \\ 
\bottomrule
\end{tabular}%
}
\label{tab:dataset_statistics}
\end{table*}

% \newpage

\subsection{GNN-AutoGL Benchmark Construction}

\label{sec:gnn_autogl_benchmark}

For the GNN-AutoGL benchmark, we use a unified template for construction. Specifically, we ask the large language model (LLM) to write code that constructs a GNN model using AutoGL based on the given prompt. Given the numerous model parameters, specifying each parameter and requiring the LLM to write corresponding model code is challenging. Therefore, we select only some parameters as requirements and evaluate whether the LLM correctly includes them. For a problem with $K$ specified parameters, if the code runs successfully, we use a Python script to assess the correctness of the $k$ specified parameters, resulting in a score of $k/K$. If the code fails to run, the score is zero. Below is an example of problems in this benchmark.

\begin{tcolorbox}[colback=gray!10, colframe=black, rounded corners, boxrule=1.5pt, fontupper=\normalsize, left=2mm, right=2mm, top=1mm, bottom=1mm]
Given the cora dataset, can you use autogl to train a $2$-layer $gcn$ model with hidden dimension as $[128, 256]$, dropout as $0.6$ and predict the accuracy ?
\end{tcolorbox}

A Python code example for solving the above problem is shown below.

\begin{tcolorbox}[colback=gray!10, colframe=black, rounded corners, boxrule=1.5pt, fontupper=\normalsize, left=2mm, right=2mm, top=1mm, bottom=1mm]
    \begin{verbatim}
import autogl
...
from autogl.datasets import 
build_dataset_from_name
cora_dataset = 
build_dataset_from_name(`cora')

...


encoder_hp = {
    `num_layers': 2,
    `hidden': [128, 256],
    `dropout': 0.6,
    `act': 'relu',
    `eps': 'false'
}

...

predicted = solver.predict_proba()
label = 
get_graph_labels(cora_dataset[0])
[get_graph_masks(cora_dataset[0], 
`test')].cpu().numpy()
print(`Test accuracy: ', 
Acc.evaluate(predicted, label))
    \end{verbatim}
    
\end{tcolorbox}

\subsection{Metric Details}
\label{sec:metrics-details}
Specifically, we extract the generated codes from model responses, and compare the execution results with the reference solutions to assess whether the model-generated results match the target answers. For the first five benchmarks~\cite{talk_like_a_graph, graphwiz, LLM4DyG, graphinstruct, nlgraph}, the evaluation scores are binary based on the exact matches between execution results and target answers. For GNN-AutoGL, we first evaluate whether the code can be executed without errors; if it cannot, it scores 0. Otherwise, we check whether the hyper-parameters (\text{e.g.,} model name, dataset name, hidden dimension, dropout) meet the requirements. Each correct hyper-parameter earns partial credit, and if all hyper-parameters are correct, the score is 1.

\section{Additional Experiments}

\begin{figure}[!ht]
    \centering
    % \vspace{-1em}
    \includegraphics[width=1.0\linewidth]{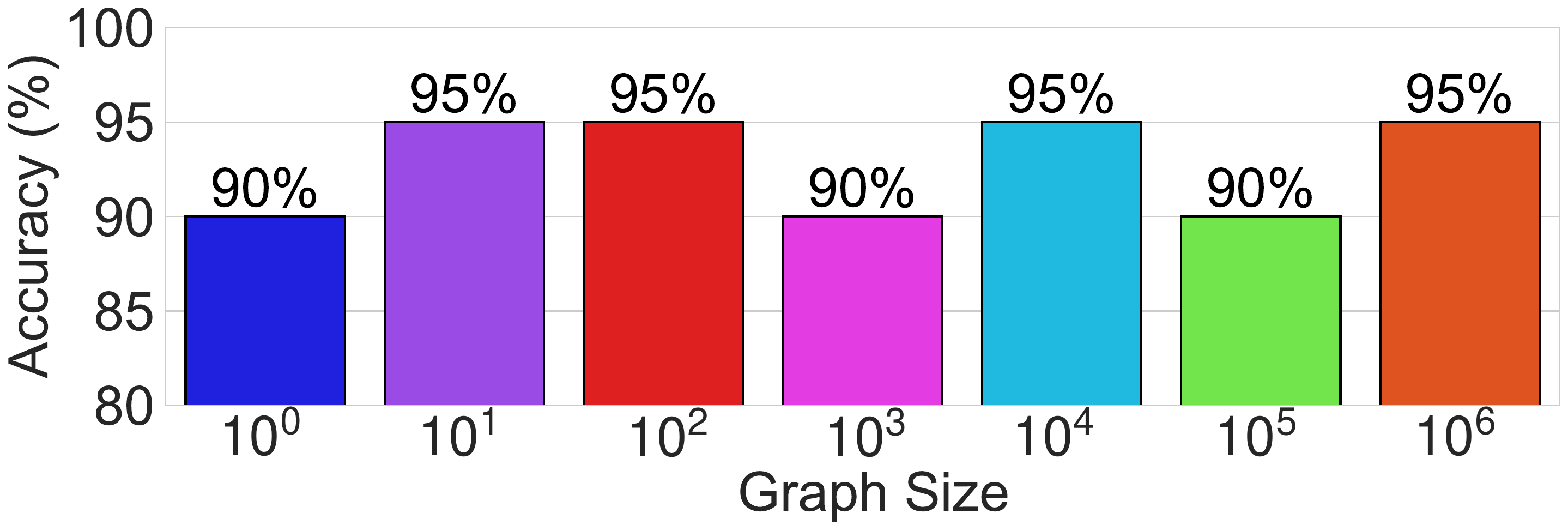}
    \vspace{-2.0em}
    \caption{Large scale graph experiments on NLGraph}
    \label{fig:large_scale_graph_on_NLGraph}
\end{figure}

\begin{table*}[htbp]
    % \vspace{-1em}
  \caption{\small Performance comparison on \textbf{hard} subsets of six graph analysis benchmarks in terms of accuracy (\%).}
  % \vspace{-1em}
  \label{tab:main-experiments-hard}
  \centering
  \rowcolors{2}{white}{gray!15}
  \resizebox{1.0\textwidth}{!}{%
  \begin{tabular}{llcccccc}
    \toprule
    Category & 
    Models & NLGraph & \makecell{Talk like a Graph} & GraphInstruct & GraphWiz & LLM4DyG & GNN-AutoGL \\
    \midrule
    % 基本类别
    % \rowcolor{rowgray1}
    {\cellcolor{white}Main}  & GPT-4o-mini & 44.9 & 47.9 & 23.2 & 28.9 & 42.6 & 3.3\\
    % \midrule
    \cellcolor{white}& SOTA & 45.0 & 93.9 & 11.7 & 70.5 & 33.9 & - \\
    % \rowcolor{rowgray2} 
    % Basic 
    % \midrule
    % \rowcolor{rowgray1} 
    % \multirow{-2}{*}{\cellcolor{white}Basic}&  \textbf{\modelname} 
    \multirow{-2}{*}{\cellcolor{white}}&  \textbf{\modelname (GPT-4o-mini)} & \textbf{98.1} & \textbf{99.2} & \textbf{73.3} & \textbf{76.3} & \textbf{98.3} & \textbf{48.5} \\
    \midrule
    \cellcolor{white} & \hspace{1em}- question agent & 69.8 & 70.0 & \underline{73.2} & 39.8 & 53.6 & 14.5\\ 
    % \rowcolor{rowgray2}
    \multirow{-2}{*}{\cellcolor{white}\makecell[l]{Input-Output\\Normalization}} & \hspace{1em}- answer agent & \underline{97.5} & \underline{97.4} & 70.0 & \underline{75.3} & \underline{96.3} & 45.8 \\
    \midrule
    % 知识检索类别
    % \rowcolor{rowgray1}
    \cellcolor{white} & \hspace{1em}- documentation & \underline{97.5} & \underline{98.9} & 69.8 & \underline{75.3} & 94.3 & \underline{46.8} \\
    % \rowcolor{rowgray2}
    \multirow{-2}{*}{\cellcolor{white}\makecell[l]{Knowledge\\ Retrieval}} & \hspace{1em}- experience & 42.3 & 98.4 & 64.0 & 62.9 & 70.6 & 4.8 \\
    % 输入输出归一化类别
    % \rowcolor{rowgray1}
    \midrule
    % \multirow{-1}{*}
    % \midrule
    % 问题解决类别
    % \rowcolor{rowgray2}
    \cellcolor{white}& \hspace{1em}- coding agent & 19.2 & 64.0 & 38.2 & 38.3 & 51.6 & - \\
    % \rowcolor{rowgray1}
    \cellcolor{white}& \hspace{1em}- retry & 40.3 & 95.4 & 71.0 & 61.0 & 85.4 & 44.3 \\
    % \rowcolor{rowgray2}
    \multirow{-3}{*}{\cellcolor{white}\makecell[l]{Problem\\Solving}} & \hspace{1em}- reasoning agent & \underline{97.5} & \underline{98.9} & 69.6 & 66.1 & 95.7 & 46.5 \\
    \bottomrule
  \end{tabular}
  }
  % \vspace{-1.5em}
\end{table*}

\subsection{Explorations on Large-scale Graphs}

The graph sizes in existing benchmarks are relatively small. To validate the effectiveness of \modelname in solving large-scale graph reasoning problems, we use NLGraph's problem generation tool to randomly create 20 problems for each graph scale of $10^1$, $10^2$, $10^3$, $10^4$, $10^5$, and $10^6$ nodes. Then \modelname (GPT-4o-mini) is employed to solve these generated problems. For graphs with at least $10^3$ nodes, \modelname is asked to load them from files via programming. The results, shown in Figure~\ref{fig:large_scale_graph_on_NLGraph}, demonstrate that our multi-agent framework can effectively handle large-scale graph reasoning tasks, maintaining 90\% accuracy for graphs containing $10^6$ nodes. 

\subsection{Ablation Study: Analysis on Hard Subsets}
To investigate \modelname's performance on challenging problems, we assess its performance on the ``hard'' subset of each benchmark. For GraphWiz, we adhere to the official partition and report results on their designated hard set. For other benchmarks, which all have their problems categorized into different types or domains, we employ a model-based approach: we identify the three categories where previous SOTA performs poorest and select problems from these categories. This selection method ensures we target the most challenging aspects of each benchmark. 

Tab.~\ref{tab:main-experiments-hard} summarizes the system's performance on these high-difficulty questions. Notably, \modelname maintains high accuracy on the Talk like a Graph hard subset (99.2\%), and shows significant improvements over SOTA for NLGraph (98.1\% v.s. 45.0\%) and LLM4DyG (98.3\% v.s. 33.9\%). While performance drops are observed for GraphWiz and GraphInstruct compared to the general set, \modelname still substantially outperforms both GPT-4o-mini and SOTA methods. The \textbf{retry} mechanism shows clear benefits for \textbf{hard} problems with a maximum improvement of 57.8\%. This indicates that hard problems require more complex codes, which cannot be written correctly in a single attempt and needs multiple tries to solve effectively. This consistent performance across diverse hard subsets underscores \modelname's robustness.

\subsection{Ablation Study: Analysis on Different Groupings}
Here we further compare with the ablated variants under different groupings.

\begin{figure*}[!ht]
    \vspace{-1em}
    \centering
    \includegraphics[width=1.0\linewidth]{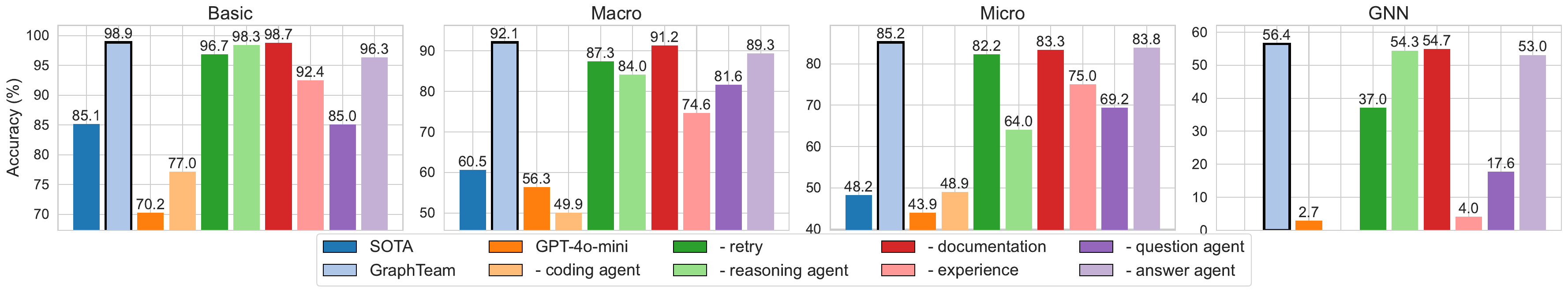}
    \vspace{-1.0em}
    \caption{Performance with respect to different task categories.}
    \label{fig:question_type_appendix}
    \vspace{-1em}
\end{figure*}

\textbf{Group by Task Categories.}
The results are shown in Figure~\ref{fig:question_type_appendix}.  For micro-level tasks, removing the reasoning agent causes a significant 21.2\% drop in performance, highlighting the complexity of these problems and the limitations of the coding agent alone. In GNN-related tasks, while the coding agent and experience are important, the question agent plays a crucial role (38.8\% impact when removed). This underscores the importance of effectively extracting key information from GNN problem statements. These findings point to areas for focused improvement, particularly in enhancing reasoning capabilities for micro-level tasks and information extraction for GNN-related problems.

\begin{figure*}[!ht]
    \centering
    \vspace{-1em}
    \includegraphics[width=1.0\linewidth]{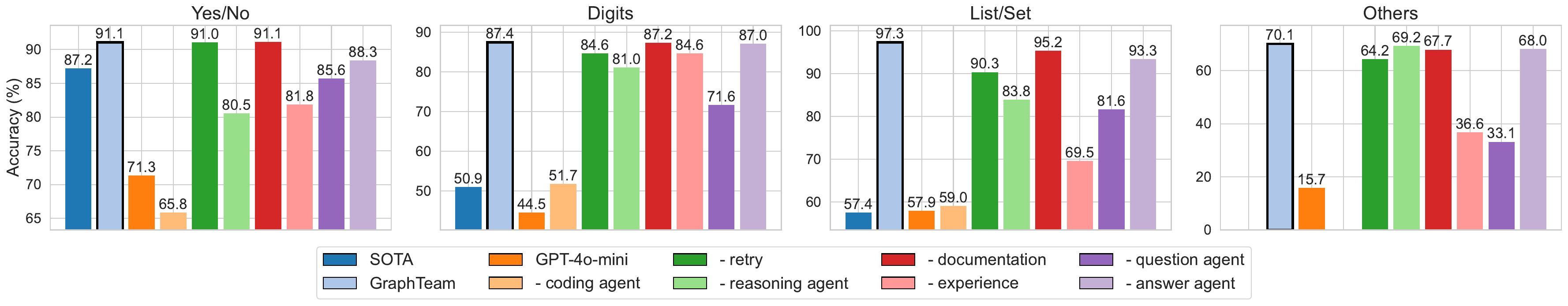}
    \vspace{-1em}
    \caption{Performance with respect to different output formats.}
    \label{fig:output_type}
    \vspace{-1em}
\end{figure*}

\textbf{Group by Output Formats.} As shown in Figure~\ref{fig:output_type}, \modelname demonstrates strong performance on yes/no, digits, and list/set outputs, achieving an accuracy of over 85\% for each of these categories. However, the system's performance drops notably for more complex output formats, with an accuracy of only 70.1\%. This decline in accuracy for the ``others'' format indicates that \modelname, particularly its answer agent, needs enhancement in handling advanced data structures and more complex output formats. Improving the system's capability to generate and structure complex outputs could significantly boost its overall effectiveness in addressing a wider range of graph analysis problems.

\subsection{Detailed Results for Each Benchmark}

In this section, we present comprehensive results for the subclasses of five benchmarks: NLGraph, Talk like a graph, GraphWiz, LLM4DyG, and GraphInstruct. We employ GPT-4o-mini with the original state-of-the-art (SOTA) results for these benchmarks. 

\subsubsection{NLGraph}

\begin{table*}[htbp]
\vspace{-1em}
\centering
\caption{Performance comparison on NLGraph benchmark in terms of accuracy (\%).}
\resizebox{\textwidth}{!}{%
\begin{tabular}{cccccccccc} % 10 列居中对齐
\toprule
\makecell[c]{} & \makecell[c]{Connect} & \makecell[c]{Cycle} & \makecell[c]{Topo. Sort} & \makecell[c]{Shortest Path} & \makecell[c]{Max. Flow} & \makecell[c]{Bipartite Graph} & \makecell[c]{Hamilton Path} & \makecell[c]{GNN} &
\makecell[c]{Overall \\ Average} \\ 
\midrule
GPT-4o-mini & 72.0 & 52.9 & 20.0 & 35.9 & 98.3 & 29.8 & 22.4 & 0 & 51.4 \\ 
Previous SOTA & \textbf{99.5} & 96.9 & 63.7 & 60.9 & 17.2 & 57.1 & 39.7 & 94.9 & 79.7 \\ 
\modelname (GPT-4o-mini) & 97.0 & \textbf{100.0} & \textbf{94.8} & \textbf{98.4} & \textbf{100.0} & \textbf{100.0} & \textbf{100.0} & \textbf{97.4} & \textbf{97.8} \\ 
\bottomrule
\end{tabular}%
}
\label{tab:nlgraph}
% \vspace{-1em}
\end{table*}

As shown in Table~\ref{tab:nlgraph}, we present detailed results across subclasses of NLGraph. \modelname (GPT-4o-mini) achieves perfect scores in multiple categories (including Cycle, Maximum Flow, Bipartite Graph, and Hamilton Path), with its lowest performance being 94.8 in Topology Sort. These results substantially exceed the previous SOTA~\cite{GUNDAM} and GPT-4o-mini baseline.

\subsubsection{Talk Like a Graph}

\begin{table*}[!ht]
\vspace{-1em}
\centering
\caption{Performance comparison on Talk like a graph benchmark in terms of accuracy (\%).}
\resizebox{\textwidth}{!}{%
\begin{tabular}{cccccccccc} % 10 列居中对齐
\toprule
\makecell[c]{} & 
\makecell[c]{Node Count} & \makecell[c]{Edge Count} & 
\makecell[c]{Edge Existence} & \makecell[c]{Node Degree} & \makecell[c]{Connected Nodes} & \makecell[c]{Cycle Check} & \makecell[c]{Shortest Path} & 
\makecell[c]{Triangle Counting} &
\makecell[c]{Overall Average} \\ 
\midrule
GPT-4o-mini & \textbf{100.0} & 45.6 & 87.0 & 54.6 & 50.2 & 92.6 & 45.6 & 14.4 & 61.3 \\ 
Previous SOTA & \textbf{100.0} & \textbf{100.0} & \textbf{96.1} & 91.7 & 98.0 & 98.0 & 97.2 & 40.5 & 90.2 \\ 
\modelname (GPT-4o-mini) & \textbf{100.0} & 99.2 & 61.0 & \textbf{98.6} & \textbf{99.2} & \textbf{99.4} & \textbf{99.1} & \textbf{99.2} & \textbf{94.5} \\ 
\bottomrule
\end{tabular}%
}
\label{tab:talk-like-a-graph}
% \vspace{-1em}
\end{table*}

As shown in Table~\ref{tab:talk-like-a-graph}, \modelname (GPT-4o-mini) achieves over 90.0\% accuracy across the vast majority of subclasses. For the edge existence task, we encounter performance limitations due to ambiguity in problem descriptions. Despite relatively weak performance on this task, \modelname (GPT-4o-mini) maintains an average performance above 90.0\%, demonstrating strong generalization capability.

\newpage

\subsubsection{GraphInstruct}

\;

\begin{table*}[!ht]
\centering
\vspace{-2.5em}
\caption{Performance comparison on GraphInstruct benchmark in terms of accuracy (\%).}
\resizebox{\textwidth}{!}{%
\begin{tabular}{cccccccccc} % 10 列居中对齐
\toprule
\makecell[c]{} & 
\makecell[c]{Neighbor} & \makecell[c]{Bipatite} & \makecell[c]{Edge} & 
\makecell[c]{Connectivity} & \makecell[c]{Degree} & \makecell[c]{DFS} & 
\makecell[c]{Predecessor} & 
\makecell[c]{Topological \\ Sort} &
\makecell[c]{Connected \\ Component} \\ 
\midrule
GPT-4o-mini & 98.0 & 89.7 & 90.7 & 92.2 & 98.0 & 89.3 & 47.4 & 59.0 & 39.3 \\ 
Previous SOTA & 99.0 & 85.0 & 84.0 & 83.0 & 81.0 & 46.0 & 41.0 & 33.0 & 83.0 \\ 
\modelname (GPT-4o-mini) & \textbf{100.0} & \textbf{99.0} & \textbf{99.3} & \textbf{100.0} & \textbf{100.0} & \textbf{95.7} & \textbf{98.0} & \textbf{98.3} & \textbf{94.7} \\ 
\midrule
\makecell[c]{} & 
\makecell[c]{Common \\ Neighbor} & \makecell[c]{Hamiltonian \\ Path} & 
\makecell[c]{Jaccard} & \makecell[c]{Shortest \\ Path} & \makecell[c]{Diameter} & \makecell[c]{PageRank} & \makecell[c]{Maximum \\ Flow} & 
\makecell[c]{MST} &
\makecell[c]{Overall \\ Average} \\ 
\midrule
GPT-4o-mini & 49.0 & 33.7 & 17.0 & 57.3 & 23.0 & 15.3 & 15.0 & 1.0 & 53.7 \\ 
Previous SOTA & 23.0 & 18.0 & 14.0 & 14.0 & 13.0 & 13.0 & 6.0 & \textbf{4.0} & 43.5 \\ 
\modelname (GPT-4o-mini) & \textbf{100.0} & \textbf{97.7} & \textbf{100.0} & \textbf{100.0} & \textbf{100.0} & \textbf{16.3} & \textbf{99.3} & 0 & \textbf{88.1} \\
\bottomrule
\end{tabular}%
}
\label{tab:graphinstruct1}
\end{table*}

As shown in Table~\ref{tab:graphinstruct1}, results show that \modelname can solve the vast majority of GraphInstruct tasks with accuracy exceeding 90.0\%, except for MST (minimum spanning tree) tasks. The poor performance on MST tasks is attributed to a fundamental limitation where current mainstream LLMs lack the required knowledge, resulting in poor performance across all baselines as well as our \modelname. In the future, we plan to enhance \modelname to address such challenges. Nevertheless, \modelname significantly outperforms previous SOTA on all other subclasses of GraphInstruct. 

\subsubsection{GraphWiz}

\begin{table*}[htbp]
\centering
\vspace{-1em}
\caption{Performance comparison on GraphWiz benchmark in terms of accuracy (\%).}
\resizebox{\textwidth}{!}{%
\begin{tabular}{ccccccccccc} % 10 列居中对齐
\toprule
\makecell[c]{} & 
\makecell[c]{Cycle} & \makecell[c]{Connect} & \makecell[c]{Bipartite} & \makecell[c]{Topology} & \makecell[c]{Shortest} & \makecell[c]{Triangle} & \makecell[c]{Flow} & 
\makecell[c]{Hamilton} &
\makecell[c]{Subgraph} &
\makecell[c]{Overall \\ Average} \\ 
\midrule
GPT-4o-mini & 82.0 & 64.8 & 67.3 & 100.0 & 8.5 & 10.3 & 19.3 & 36.3 & 40.3 & 47.6 \\ 
Previous SOTA & 89.0 & 82.5 & 84.8 & 46.8 & 24.0 & 52.8 & 43.5 & \textbf{81.5} & 77.3 & 65.0 \\ 
\modelname (GPT-4o-mini) & \textbf{100.0} & \textbf{86.5} & \textbf{96.3} & \textbf{100.0} & \textbf{95.0} & \textbf{94.0} & \textbf{94.0} & 32.5 & \textbf{99.3} & \textbf{88.6} \\ 
\bottomrule
\end{tabular}%
}
\label{tab:graphwiz}
\end{table*}

The results are shown in Table~\ref{tab:graphwiz}. Although both GPT-4o-mini and \modelname (GPT-4o-mini) can solve the Hamilton problem conceptually, their performance is substantially impacted by the inability to match the complex output format required by the evaluation script of GraphWiz. The previous SOTA model~\cite{graphwiz} is fine-tuned on extensive datasets, and thus able to meet the required output format. Nevertheless, \modelname achieves an overall accuracy of 88.1\% on GraphWiz, demonstrating strong graph reasoning capabilities.

\subsubsection{LLM4DyG}

\begin{table*}[htbp]
\centering
\vspace{-1em}
\caption{Performance comparison on LLM4DyG benchmark in terms of accuracy (\%).}
\resizebox{\textwidth}{!}{%
\begin{tabular}{ccccccccccc} % 10 列居中对齐
\toprule
\makecell[c]{} & 
\makecell[c]{When \\ Link} & \makecell[c]{When \\ Connect} & \makecell[c]{When \\ Tclosure} & \makecell[c]{Neighbor \\
at Time} & \makecell[c]{Neighbor
\\ in Periods} & \makecell[c]{Check
 \\ Tclosure} & \makecell[c]{Check
\\ Tpath} & 
\makecell[c]{Find
\\ Tpath} &
\makecell[c]{Sort
\\ Edge} &
\makecell[c]{Overall \\ Average} \\ 
\midrule
GPT-4o-mini & 58.0 & 67.3 & 87.7 & 33.3 & 44.7 & 77.3 & 57.7 & 46.7 & 49.7 & 58.0 \\ 
Previous SOTA & 47.3 & 64.1 & 63.0 & 45.9 & 20.0 & 66.7 & 59.8 & 80.8 & 35.8 & 53.7 \\ 
\modelname (GPT-4o-mini) & \textbf{96.3} & \textbf{94.3} & \textbf{100.0} & \textbf{95.3} & \textbf{100.0} & \textbf{99.7} & \textbf{87.0} & \textbf{81.0} & \textbf{99.7} & \textbf{94.8} \\ 
\bottomrule
\end{tabular}%
}
\label{tab:llm4dyg}
\end{table*}

As shown in Table~\ref{tab:llm4dyg}, \modelname (GPT-4o-mini) demonstrates exceptional performance, achieving over 90.0\% accuracy on multiple tasks (including 100.0\% accuracy on When T-closure and Neighbor in Periods). Even its lowest performance, on Find Tpath, reaches 81.0\% accuracy, significantly outperforming previous SOTA.

%%%%%%%%%%%%%%%%%%%%%%%%%%%%%%%%%%%%%%%%%%%%%%%%%%%%%%%%%%%%

\end{document}